\documentclass{article}

\usepackage{arxiv}

\usepackage[utf8]{inputenc} 
\usepackage[T1]{fontenc}    
\usepackage{hyperref}       
\usepackage{url}            
\usepackage{booktabs}       
\usepackage{amsfonts}       
\usepackage{nicefrac}       
\usepackage{microtype}      
\usepackage{lipsum}

\usepackage{graphicx, subfigure, multirow, array, amsmath}
\usepackage[super]{nth}

\newcommand{\rehl}[1] {#1}

\graphicspath{{fig/}}

\title{Context awareness and embedding for biomedical event extraction}

\author{
  Shankai Yan\\
  Department of Computer Science\\
  City University of Hong Kong\\
  Hong Kong SAR\\
  \texttt{sk.yan@my.cityu.edu.hk}\\
   \And
  Ka-Chun Wong\\
  Department of Computer Science\\
  City University of Hong Kong\\
  Hong Kong SAR\\
  \texttt{kc.w@cityu.edu.hk} \\
}

\begin{document}
\maketitle

\begin{abstract}
\textbf{Motivation:} Biomedical event detection is fundamental for information extraction in molecular biology and biomedical research. The detected events form the central basis for comprehensive biomedical knowledge fusion, facilitating the digestion of massive information influx from literature. Limited by the feature context, the existing event detection models are mostly applicable for a single task. A general and scalable computational model is desiderated for biomedical knowledge management.\\
\textbf{Results:} We consider and propose a bottom-up detection framework to identify the events from recognized arguments. To capture the relations between the arguments, we trained a bi-directional Long Short-Term Memory (LSTM) network to model their context embedding. Leveraging the compositional \rehl{attributes}, we further derived the candidate samples for training event classifiers. We built our models on the datasets from BioNLP Shared Task for evaluations. Our method achieved the average F-scores of 0.81 and 0.92 on BioNLPST-BGI and BioNLPST-BB datasets respectively. Comparing with 7 state-of-the-art methods, our method nearly doubled the existing F-score performance (0.92 vs 0.56) on the BioNLPST-BB dataset. Case studies were conducted to reveal the underlying reasons.\\
\textbf{Availability:} https://github.com/cskyan/evntextrc\\
\end{abstract}


\section{Introduction}

The unbridled growth of publications in biomedical literature databases offers a great opportunity for researchers to stand on the shoulders of giants for cutting-edge advancements. Nonetheless, it is also a challenge to digest the extensive information from the huge volume of textual data. Information Extraction (IE) is an effective approach to summarize the knowledge into expressive forms for management and comprehension; it can be integrated with other knowledge resources for innovative discovery \cite{Rebholz-Schuhmann2012}. Examples include protein-protein interactions \cite{Mallory2015}, drug-drug interaction \cite{Zhao2016}, causal relationships between biological entities \cite{Perfetto2015}, and other topic-oriented association mining systems \cite{Lim2016, Canada2017}.

Over the past decades, considerable efforts have been devoted towards rule-based \cite{Bui2012} and trigger-based \cite{Bjorne2010, Bjorne2011} detection methods for biomedical event extraction from PubMed abstracts \cite{Ananiadou2010}. \rehl{In general, trigger detection dominates the whole prediction process which performance will greatly affect the final event detection \cite{Pyysalo2012}. Trigger identification method has been well-studied and improved. The latest trigger-based approach using deep neural network has shown its strength in general event extraction tasks \cite{Nguyen2016}.} Combining with lexical and semantic features, word embedding \cite{Mikolov2013} is proposed to build an advanced trigger classifier \cite{Zhou2014}. Nevertheless, trigger detection is a multi-class classification problem with limited annotation labels. The well-known datasets from BioNLP Shared Task (BioNLPST) include BioNLP'09 \cite{Kim2009}, BioNLP'11 \cite{Kim2011}, BioNLP'13 \cite{Nedellec2013}, and BioNLP'16 \cite{Nedellec2016}. The trigger-based methods are based on the dependency parse tree and character n-grams. The dependency parser in natural language processing (NLP) is well-studied \cite{Nivre2016} and has been developed from empirical techniques to neural network models \cite{Chen2014}. However, there is a performance deviation from the traditional applications when applying to biomedical literature due to the contextual variations. The parser that was developed specifically for biomedical text mining (BioNLP) such as \rehl{McCCJ \cite{McClosky2008}} is necessary for biomedical information extraction \cite{Luo2017}. Bi-directional LSTM has been applied to medical event detection in clinical records \cite{Jagannatha2016}. \rehl{Nonetheless, its events are binary relations which are very different from the complex events in BioNLPST.}

\rehl{One of the major concerns behind this is that the prediction errors would propagate along the whole pipeline.} The training data for trigger detection is quite limited because the ground truth labels are not even given in the BioNLP Shared Task datasets. \rehl{In addition, the training samples are not easily selected manually.} Consequently, it becomes an unbalanced multi-class classification problem which is the main barrier to performance improvement in the subsequent biomedical text mining tasks.

\rehl{In this study, we proposed a novel method to detect biomedical events using a different strategy. We do not need the annotation of triggers and the cumbersome dependency parsing for each sentence. We aspire to model the context embedding for each argument. The argument embeddings are adopted to detect directed relations. The proposed neural network model is applicable to general event extraction, thanks to the universality of the underlying neural language models \cite{Bengio2003}. Our method is specially designed for biomedical event extraction while keeping replaceable components (e.g. pre-trained word embedding) for general event extraction tasks. The remainder of this paper is organized in the following order. Firstly, we briefly introduce the datasets and indicate the defectiveness of the existing approaches. Next, we sketch out the framework of our approach and then elaborate the procedures in detail. After that, we evaluate our method and make a comprehensive comparison with other approaches on the BioNLP Shared Task dataset. Then, we demonstrate the effectiveness of our method by investigating the underlying reasons through experiments.}

\section{Datasets}

\begin{table}[!ht]
  \caption{Statistics of the events for two tasks in BioNLP Shared Task \label{tab-evntstat}} \centering{\renewcommand{\arraystretch}{1.4}
\begin{tabular}{@{}m{\dimexpr 0.2\linewidth-2\tabcolsep}m{\dimexpr 0.3\linewidth-2\tabcolsep}m{\dimexpr 0.33\linewidth-2\tabcolsep}cc@{}}
\hline
Task & Event Type & Arguments & \shortstack{Train\\-ing \\Set} & \shortstack{Develo\\-pment \\Set} \\
\hline
\multirow{10}{*}{\shortstack{BioNLP \\Shared \\Task 2011 \\- Bacteria \\Gene \\Interactions}} & ActionTarget & Action-\textgreater{}Target & 108 & 18 \\
 & Interaction & Agent-\textgreater{}Target & 126 & 18 \\
 & PromoterDependence & Promoter-\textgreater{}Protein & 32 & / \\
 & PromoterOf & Promoter-\textgreater{}Gene & 36 & / \\
 & RegulonDependence & Regulon-\textgreater{}Target & 11 & / \\
 & RegulonMember & Regulon-\textgreater{}Member & 15 & / \\
 & SiteOf & Site-\textgreater{}Entity & 17 & / \\
 & TranscriptionBy & Transcription-\textgreater{}Agent & 25 & 3 \\
 & TranscriptionFrom & Transcription-\textgreater{}Site & 14 & / \\
\hline
\shortstack{BioNLP \\Shared \\Task 2016 \\- Bacteria \\Biotopes} & Lives\_In & Bacteria-\textgreater{}Location & 327 & 223 \\
\hline
  	\end{tabular}}{}
\end{table}

\rehl{In order to ensure fair comparisons among different approaches, we adopted two datasets from the BioNLP Shared Task with 1 (BioNLPST-BB) and 9 (BioNLPST-BGI) event type(s). They consist of the events of bacteria localization and the genetic processes concerning the bacterium Bacillus subtilis respectively.} We aim to measure how the performance change with different event types for model generalization estimation. The development set is initially used to validate the prediction model or tune the hyper-parameters. However, it only contains 3 out of 10 event types in BioNLPST-BGI. Therefore, we combine the training set and the development set as a single annotated dataset for each task. As shown in Table~\ref{tab-evntstat}, the event types are extremely imbalanced in BioNLPST-BGI; it means that the event detection is an imbalanced multi-class classification problem.

The events come from the sentences of PubMed abstracts and the biological entities are annotated by curators or name entity recognition (NER) tools. \rehl{The objective of event detection is to predict the relationships among the pre-annotated or recognized entities. For example, the sentence ``We now report that the purified product of gerE (GerE) is a DNA-binding protein that adheres to the promoters for cotB and cotC.'' has totally 6 pre-annotated entities, ``T1:purified product of gerE'', ``T2:GerE'', ``T3:DNA-binding protein'', ``T4:promoters'', ``T5:cotB'', ``T6:cotC''. It contains two ``PromoterOf'' events (E1:promoters-\textgreater{}cotB; E2:promoters-\textgreater{}cotC) and two ``Interaction'' events (E3:GerE-\textgreater{}cotB; E4:GerE-\textgreater{}cotC). The events are different from the traditional binary relations (e.g. gene-gene interaction) due to the difficulty of recognizing their directions and the diversity of the entity types as well as the event types. Under the context of knowledge graph topology, our prediction is a directed edge with a specific type instead of a plain binary relation. The before-mentioned example can be used to construct a directed graph consists of 6 nodes (entities) and 4 edges (events). We directly adopted the tokenization and NER results (e.g. ``T1:Protein'', ``T2:Protein'', ``T3:ProteinFamily'', ``T4:Promoter'', ``T5:Gene'', and ``T6:Gene'') from the annotated datasets.}

\begin{table}[!ht]
  \caption{Statistics of the arguments for two tasks in BioNLP Shared Task \label{tab-argstat}} \centering {
\begin{tabular}{@{}m{\dimexpr 0.33\linewidth-2\tabcolsep}m{\dimexpr 0.25\linewidth-2\tabcolsep}cc@{}}
\hline
Task & Argument Type & Training Set & Development Set \\
\hline
\multirow{10}{*}{\shortstack{BioNLP Shared \\Task 2011 - \\Bacteria Gene \\Interactions}} & Action & 92 & 16 \\
 & Agent & 125 & 15 \\
 & Entity & 15 & / \\
 & Gene & 36 & 3 \\
 & Member & 15 & / \\
 & Promoter & 38 & / \\
 & Protein & 29 & / \\
 & Regulon & 10 & / \\
 & Site & 29 & / \\
 & Target & 185 & 21 \\
 & Transcription & 31 & 3 \\
 \hline
\multirow{2}{*}{\shortstack{BioNLP Shared Task \\2016 - Bacteria Biotopes}} & Bacteria & 168 & 118 \\
 & Location & 260 & 184 \\
\hline
  	\end{tabular}}{}
\end{table}

Besides the event annotations (e.g. E1:T4-\textgreater{}T6, E2:T4-\textgreater{}T5, E3:T2-\textgreater{}T5, E4:T2-\textgreater{}T6), the argument labels (e.g. ``T1:Protein'', ``T2:Protein'', ``T3:ProteinFamily'', ``T4:Promoter'', ``T5:Gene'', ``T6:Gene'') within each event type are also used in our method. Table~\ref{tab-argstat} shows the summary of the argument numbers in each task. It is obvious that the labels for the argument types are also imbalanced. The arguments are all annotated on the recognized entities. Therefore, we assume that the error rate of the entity recognition is very low, and we can consider it as known information.

The triggers used in most of the existing approaches are not officially released in the datasets and they are manually annotated by the researchers. However, those trigger words vary across different tasks; it heavily requires manual preprocessing. Furthermore, the classification errors in the trigger detectors can propagate to the argument detection and event detection. \rehl{The nonexistence of trigger words does not affirm none events since different authors may have different styles of writing and the triggers are not guaranteed to appear in the sentence.} Therefore, we do not use any trigger-based method in our study. Instead, the context of the arguments within each event is considered while constructing features.

\section{Methodology}

\subsection{An overview of the event detection framework}

\begin{figure*}[!tpb]
  \resizebox{\textwidth}{!}{\centering\includegraphics{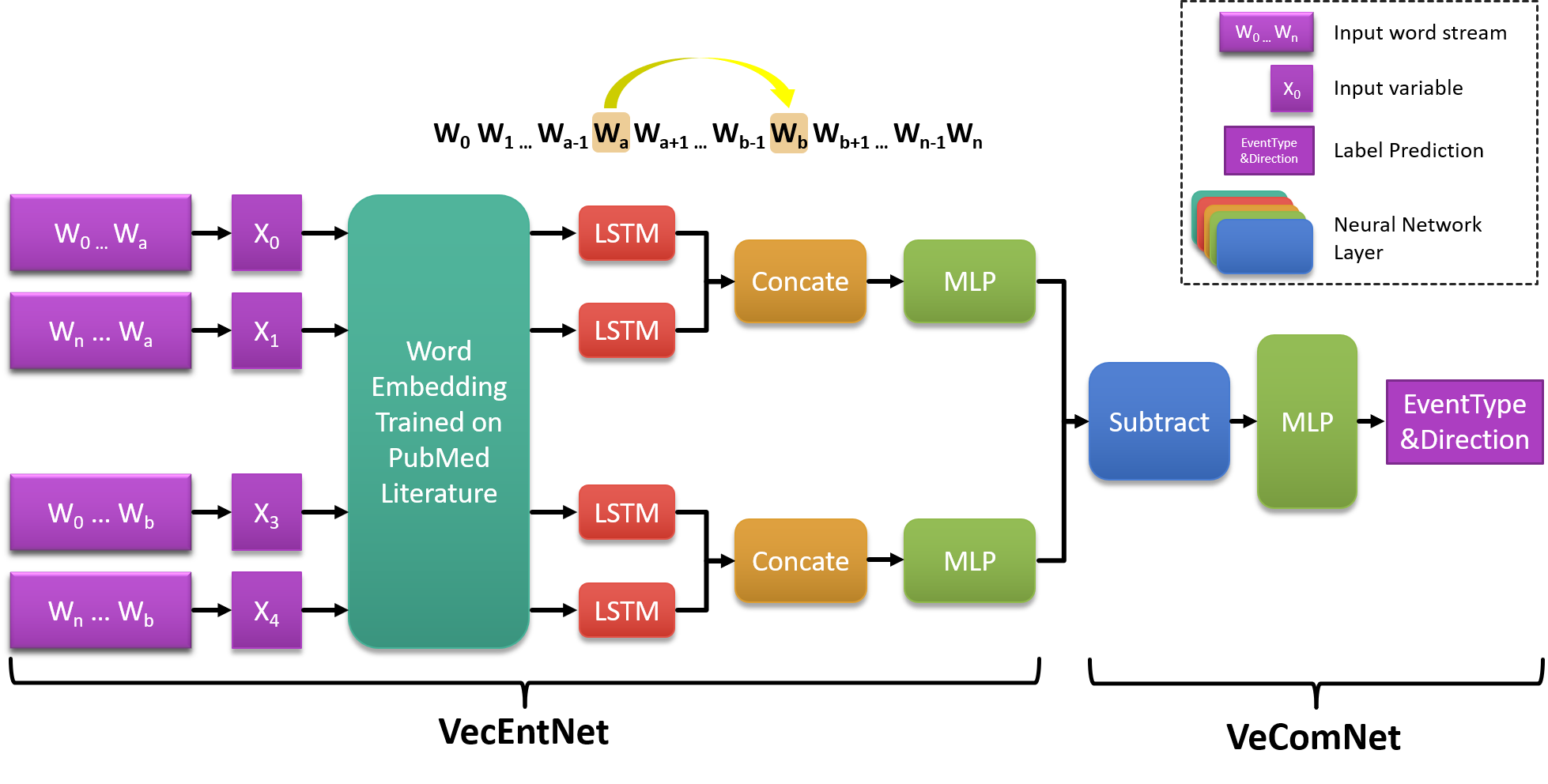}}
  \caption{Overview of the neural network architecture for argument embedding and event detection. VecEntNet is trained for argument detection using the argument annotations in the training set. The parameters of VecEntNet is then fixed and the hidden layer of the MLP in VecEntNet is used as the input of VeComNet. VeComNet is trained for directed event detection using the event annotations in the training set. The testing data is passed to VecEntNet for generating argument embedding which is put into VeComNet for event prediction.} \label{fig-wf}
\end{figure*}

The overall workflow of our proposed event detection method is shown in Figure~\ref{fig-wf}. We take the tokenized words in the dataset as input and transform them into word vectors that trained on the PubMed literature. For each event argument $W_{a}$ and $W_{b}$, we input the stream of words on both sides of them to a bi-directional LSTM for constructing the context embeddings \cite{Melamud2016} of arguments. We train the context embedding model (VecEntNet) using the annotations of arguments in each task. The context embeddings are further adopted to train the event detection model (VeComNet) for detecting event type and direction.

\subsection{Word embedding}

To construct robust features for argument recognition, we use the distributed representations of words in a sentence instead of the traditional N-gram features \cite{Mikolov2013}. The adopted word vectors are pre-trained on a corpus of 10,876,004 biomedical abstracts from PubMed, which contains 1,701,632 distinct words and 200 dimensions \cite{Kosmopoulos2015}. The training is actually a transformation from the one hot encoding of the words to a continuous space with dimension reduction. Such unsupervised training on a large corpus captures the general features of each word and help prevent over-fitting for downstream tasks.

\subsection{Bi-directional LSTM}

LSTM \cite{Gers1999} is a recurrent neural network (RNN) cell that can be trained to decide which information should be forgotten or kept. Bi-directional LSTM (BLSTM) is broadly utilized for NLP tasks to learn contextual representations from phrases or sentences \cite{Melamud2016}. Therefore, we use the words surrounding the recognized entities to train the contextual representations. As shown in Figure~\ref{fig-wf}, $W_{a}$ and $W_{b}$ are recognized as two biological entities which can be a word or a phrase. The word embedding sequences $W_{0}...W_{a}$ and $W_{n}...W{a}$ are extracted from two directions as the inputs of a bi-directional LSTM. In practice, we set up a window size $u$ to normalize the sizes of two word-sequences and use a dummy word to pad the sequence with length less than $u$. The inputs are then modified as followed.

\begin{align} \label{eq-input}
\begin{split}
\vec{x_{0}} &= <X_{0}, X_{1}> = <W_{a-u:a}, W_{a+u:a}> \\
\vec{x_{1}} &= <X_{2}, X_{3}> = <W_{b-u:b}, W_{b+u:b}>
\end{split}
\end{align}

\noindent \sloppy where $\vec{x_{i}}$ stands for the surrounding words of entity $i$, $W_{a:b}$ is the sequence of word embeddings from the $a^{th}$ to $b^{th}$ word. We adopt a closed boundary strategy to construct the contextual word sequences because the named entities itself may contain useful information to distinguish the argument context. As for the example mentioned in section 2, the word ``promoters'' itself indicates that it is probably an ``Agent'' argument in the event of ``PromoterOf'', since it is a general word that is also applicable to other entities. In contrast, the words ``cotB'' and ``cotC'' have no contributions to the context modeling, which will be forgotten in BLSTM. Given the window size $u=3$, the inputs for event promoters-\textgreater{}cotB are $\vec{x_{0}}=[adheres, to, the, promoters; and, cotB, for, promoters]; \vec{x_{1}}=[the, promoters, for, cotB; DummyWord, cotC, and, cotB]$. \rehl{The word streams are then input into $LSTM$ cells. In contrast, the output of $BLSTM$ is the concatenation output vector of the left-to-right $LSTM$ and the right-to-left $LSTM$. In the above-mentioned example, the outputs of the $BLSTM$ layers are represented as $LSTM([adheres, to, the, promoters])$ concatenations with $LSTM([and, cotB, for, promoters])$ and $LSTM([the, promoters, for, cotB])$ concatenations with $LSTM([PADDING, cotC, and, cotB])$, where $LSTM([\ldots])$ is the last output of the $LSTM$ layer.}

\begin{align} \label{eq-lstm}
\begin{split}
BLSTM(X_{0}, X_{1}) &= LSTM(W_{a-u:a}) \oplus LSTM(W_{a+u:a}) \\
BLSTM(X_{2}, X_{3}) &= LSTM(W_{b-u:b}) \oplus LSTM(W_{b+u:b})
\end{split}
\end{align}

\subsection{Argument embedding}

We use Multi-Layer Perceptron (MLP) to train the argument classification model. As observed in Table~\ref{tab-argstat}, the skewed label distribution is a challenge for argument identification. We separate this multi-label classification problem into several binary classification problems under the one-vs-all strategy. Then we train each argument classifier separately using the estimator formulated in Equation~\ref{eq-mlp}. We use Dropout layer \cite{Srivastava2014} with drop out rate 0.2 in MLP as regularization to prevent over-fitting.

\begin{align} \label{eq-mlp}
\hat{y} &= MLP(\vec{x}) = Sigmoid(F_{2}(Tanh(F_{1}(BLSTM(\vec{x})))))
\end{align}

\noindent where $F_{i}(x)=A_{i}x + b_{i}$ is a fully connected layer in MLP, $Tanh$ is the hyperbolic tangent activation function, and $Sigmoid$ is the activation function of the last layer of MLP. To tackle the imbalanced problem, we first estimate the distribution of the binary labels from the training dataset, and then use weighted binary cross-entropy in Equation~\ref{eq-loss} as the loss function to optimize the neural network model.

\begin{align} \label{eq-loss}
loss &= -\sum_{i}(zy_{i}\log\hat{y}_{i}+(1-z)(1-y_{i})\log(1-\hat{y}_{i}))
\end{align}

\noindent \rehl{where $y_{i}$ is the true label of sample $i$ and $z$ represents the weight of positive class. The class weight $z$ is estimated as $1-\frac{n}{N}$ with the number of positive class $n$ and the total number of the samples $N$.}

After VecEntNet is trained, we are able to extract the argument embedding $R$ from the first layer of the MLP (Equation~\ref{eq-argvec}) for event detection. In particular, the triggers are actually embedded in the context of each argument. The trigger information, as well as their relations to the arguments, is encoded into the argument embedding which is used for event detection.

\begin{align} \label{eq-argvec}
R(\vec{x}) &= F_{1}(BLSTM(\vec{x}))
\end{align}

For the classifier of an event type $e_{i}: <arg_{s}, arg_{t}>$, we take the input $\vec{x}^{*}$ as the concatenation of both argument embeddings for each recognized entity of candidate pairs within one sentence (Equation~\ref{eq-evnt-input}). Since we are not aware of the true argument type for each entity, we use both embedding types with different orders for the entity pairs.

\begin{align} \label{eq-evnt-input}
\vec{x}^{*} &= <R_{arg_{s}}(\vec{x_{0}}) \oplus R_{arg_{t}}(\vec{x_{0}}), R_{arg_{t}}(\vec{x_{1}}) \oplus R_{arg_{s}}(\vec{x_{1}})>
\end{align}

VeComNet is designed for detecting the event types as well as the event direction of a candidate pair of recognized entities. To be consistent, we also build the multi-class classifiers under the one-vs-all strategy for event detection. For an event type $e_{i}$, we encode the direction $arg_{s} \longrightarrow arg_{t}$ as 1 and others as 0. As a result, the label for a directed event type has two bits, in which one bit encodes the existence of this event type and another one encodes the direction. Therefore, the binary classification problem for each event type is transformed into a multi-label classification problem.

Similar to word vector, argument vector also possess the compositional attribute. To reflect the direction from the model, we use a subtract layer to combine the two input vectors as $VeCom(\vec{x}^{*})$ (Equation~\ref{eq-vecom}) and use it to predict the direction. \rehl{The subtraction of the two argument vectors can be regarded as the multiplication of the concatenation of them and a factor matrix $\begin{bmatrix} I \\ -I \end{bmatrix}$, where $I$ denotes the identity matrix. We explicitly multiply this factor matrix to conduct the vector composition before proceeding to the fully connected layer. In addition, the subtraction layer can decrease the number of neurons in the MLP, and thus its model generalization.} As for the existence, we take the $L^{1}-Norm$ of $VeCom(\vec{x}^{*})$ as input to another MLP for existence prediction.

\begin{align} \label{eq-vecom}
\begin{split}
VeCom(\vec{x}^{*}) = {} & R_{arg_{s}}(\vec{x_{0}}) \oplus R_{arg_{t}}(\vec{x_{0}})\\
{} & - R_{arg_{t}}(\vec{x_{1}}) \oplus R_{arg_{s}}(\vec{x_{1}})
\end{split}
\end{align}

\noindent The resultant directed event estimator is demonstrated in Equations~\ref{eq-evnt-mlp} and ~\ref{eq-evnt-dir} representing the existence and direction respectively.

\begin{align} \label{eq-evnt-mlp}
\begin{split}
\hat{y}^{*} &= MLP^{*}(Abs(VeCom(\vec{x}^{*}))) \\
& = Sigmoid(F_{2}^{*}(ReLU(F_{1}^{*}(Abs(VeCom(\vec{x}^{*}))))))
\end{split}
\end{align}

\begin{align} \label{eq-evnt-dir}
\begin{split}
\hat{y}' &= MLP'(VeCom(\vec{x}^{*})) \\
& = Sigmoid(F_{2}'(ReLU(F_{1}'(VeCom(\vec{x}^{*})))))
\end{split}
\end{align}

\noindent where $F_{i}^{*}(x)=A_{i}^{*}x + b_{i}^{*}$ and $F_{i}'(x)=A_{i}'x + b_{i}'$ are fully connected layers, $Abs$ is a layer for absolute value calculation, $ReLU$ is the Rectified Linear Unit activation function. Binary cross-entropy is adopted as loss function and Stochastic Gradient Descent (SGD) is adopted as the optimizer to train the classifiers for each event type.

\section{Results}

The training set and development set are combined to form an annotated dataset. We evaluate our method under 10-fold cross-validation. For the arguments or events in BioNLPST-BGI with less than 20 data instances, we change to 5-fold cross-validation to ensure that the testing set would not have less than 2 classes. To ensure the training quality of those few labels, we randomly duplicate the samples in the training set so that the binary class ratio is bounded by 5. Only the training samples are duplicated when training the argument embedding. The testing samples are neither duplicated nor used in argument embedding. We trained our models on a Linux machine equipped with a 32-core CPU and 32GB RAM. The hyper-parameters used in the experiments are summarized in Tables~\ref{tab-vecent-hyparam} and ~\ref{tab-vecom-hyparam}. \rehl{Parameter analysis is also conducted to show the robustness of our method. The results shown in the Supplementary indicate that our method are not sensitive to the hyper-parameters.}

\begin{table}[!ht]
\parbox{.5\linewidth}{
  \caption{Hyper-parameters used in VecEntNet \label{tab-vecent-hyparam}} \centering {
\begin{tabular}{@{}ll@{}}
\hline
\multicolumn{2}{c}{VecEntNet}  \\
\hline
context window size      & 10  \\
LSTM hidden/output units & 128 \\
MLP input units          & 256 \\
MLP hidden units         & 128 \\
Batch size               & 32  \\
Epoch                    & 10  \\
\hline
  	\end{tabular}}{}}
\hfill
\parbox{.5\linewidth}{
  \caption{{Hyper-paramet\newline-ers used in VeComNet} \label{tab-vecom-hyparam}} \centering {
\begin{tabular}{@{}ll@{}}
\hline
\multicolumn{2}{c}{VeComNet}   \\
\hline
MLP input units          & 128 \\
MLP hidden units         & 64  \\
Batch size               & 32  \\
Epoch                    & 10  \\
\hline
  	\end{tabular}}{}}
\end{table}

\subsection{Performance of VecEntNet and VeComNet during training}

\begin{figure*}[!tpb]
  \resizebox{\textwidth}{!}{\centering
    \subfigure[]{\includegraphics[width=.33\textwidth]{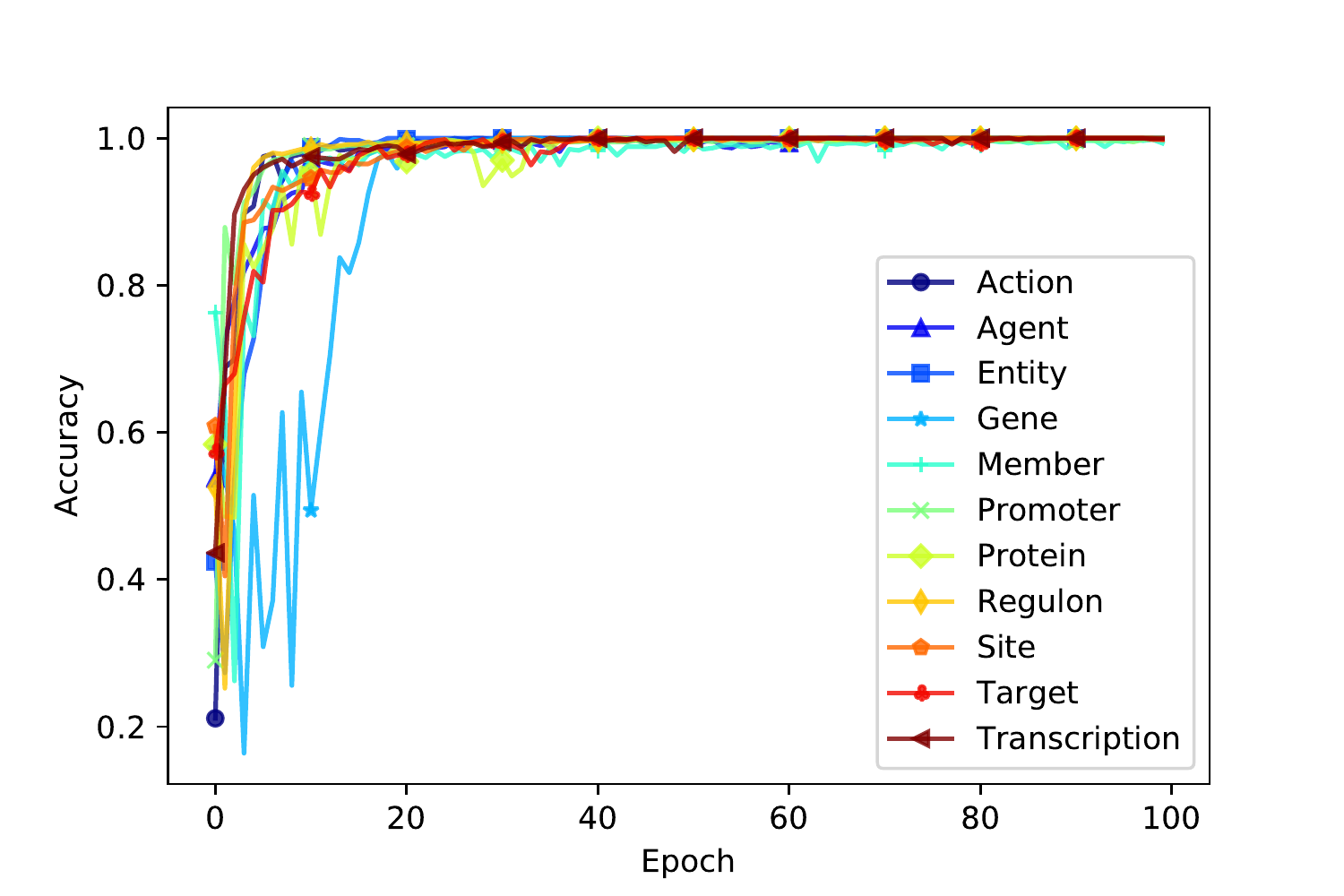}}
    \subfigure[]{\includegraphics[width=.33\textwidth]{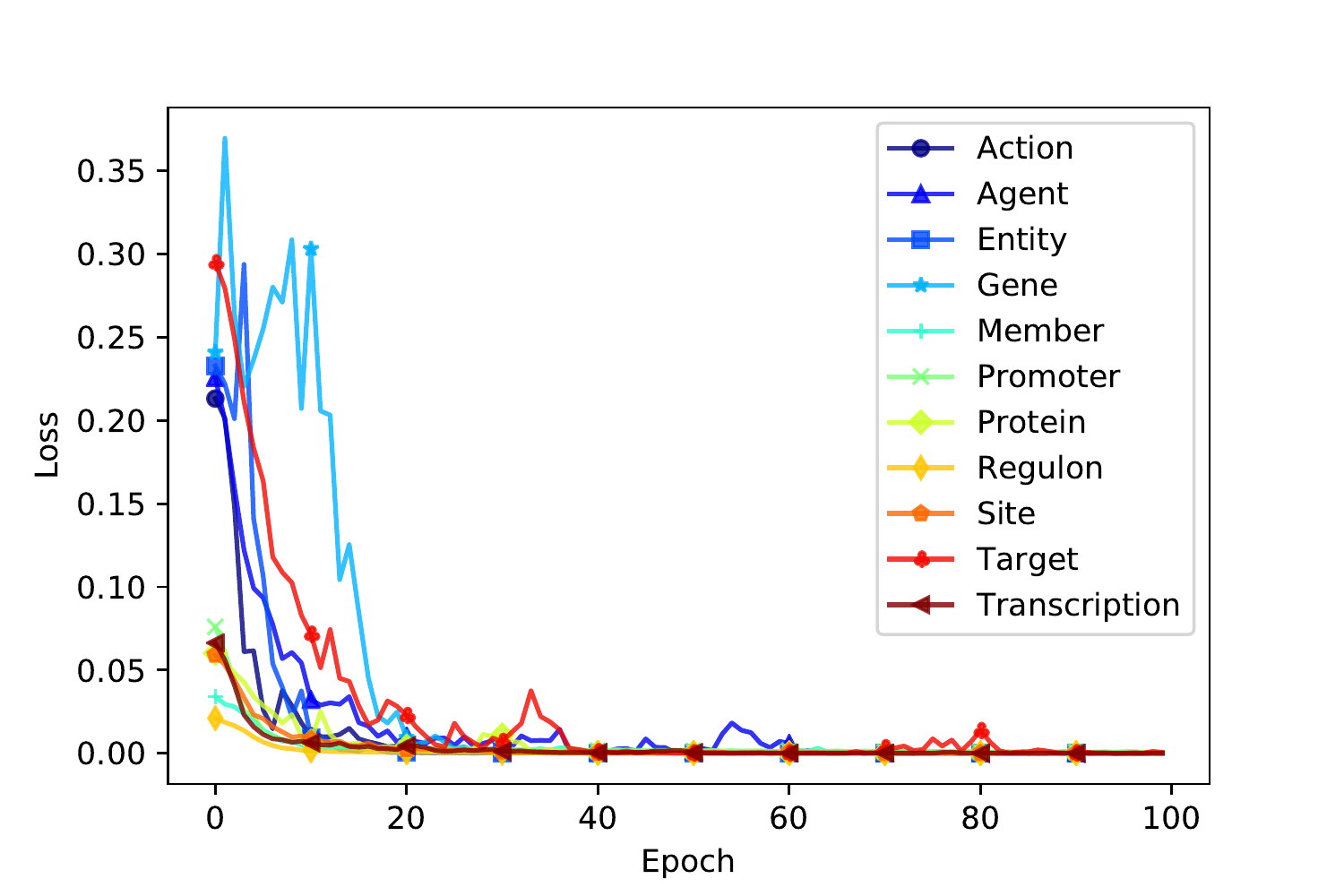}}
    \subfigure[]{\includegraphics[width=.33\textwidth]{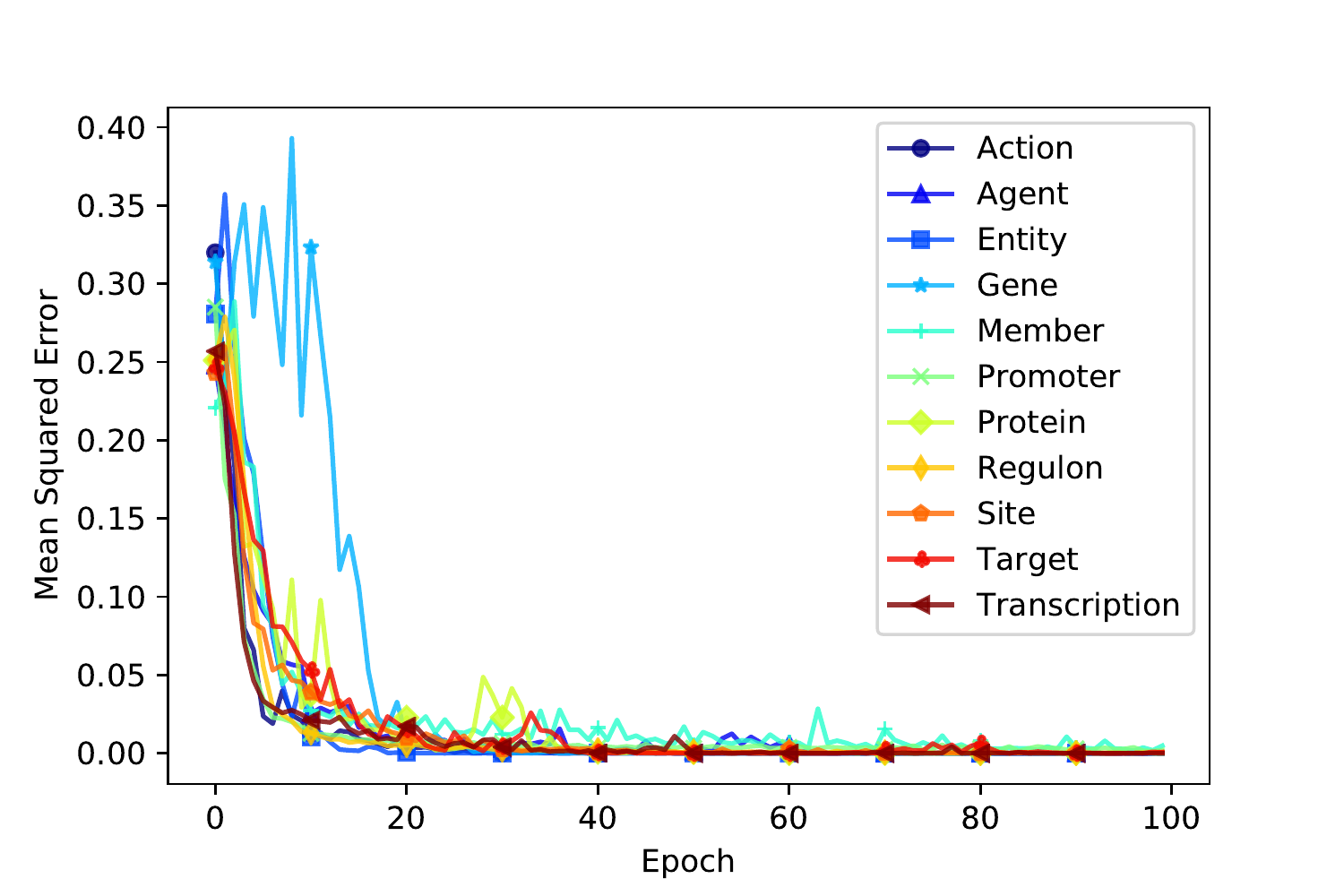}}
  }
  \caption{Performance of VecEntNet on BioNLPST-BGI dataset over training and testing process. (a) Model Accuracy. (b) Model Loss. (c) Mean Squared Error.} \label{fig-bgi2011-ent-model-perf}
\end{figure*}

\begin{figure*}[!tpb]
  \resizebox{\textwidth}{!}{\centering
    \subfigure[]{\includegraphics[width=.33\textwidth]{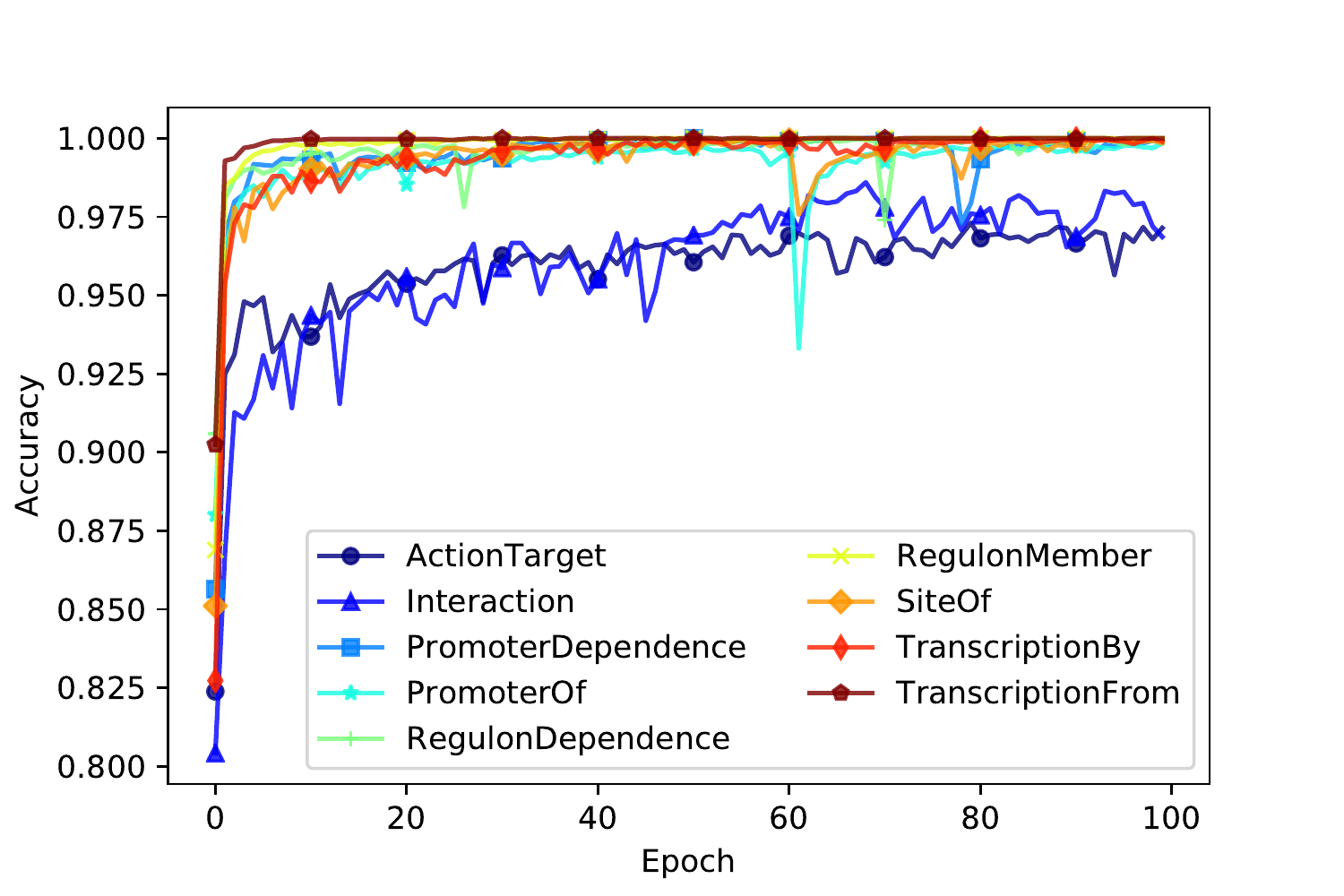}}
    \subfigure[]{\includegraphics[width=.33\textwidth]{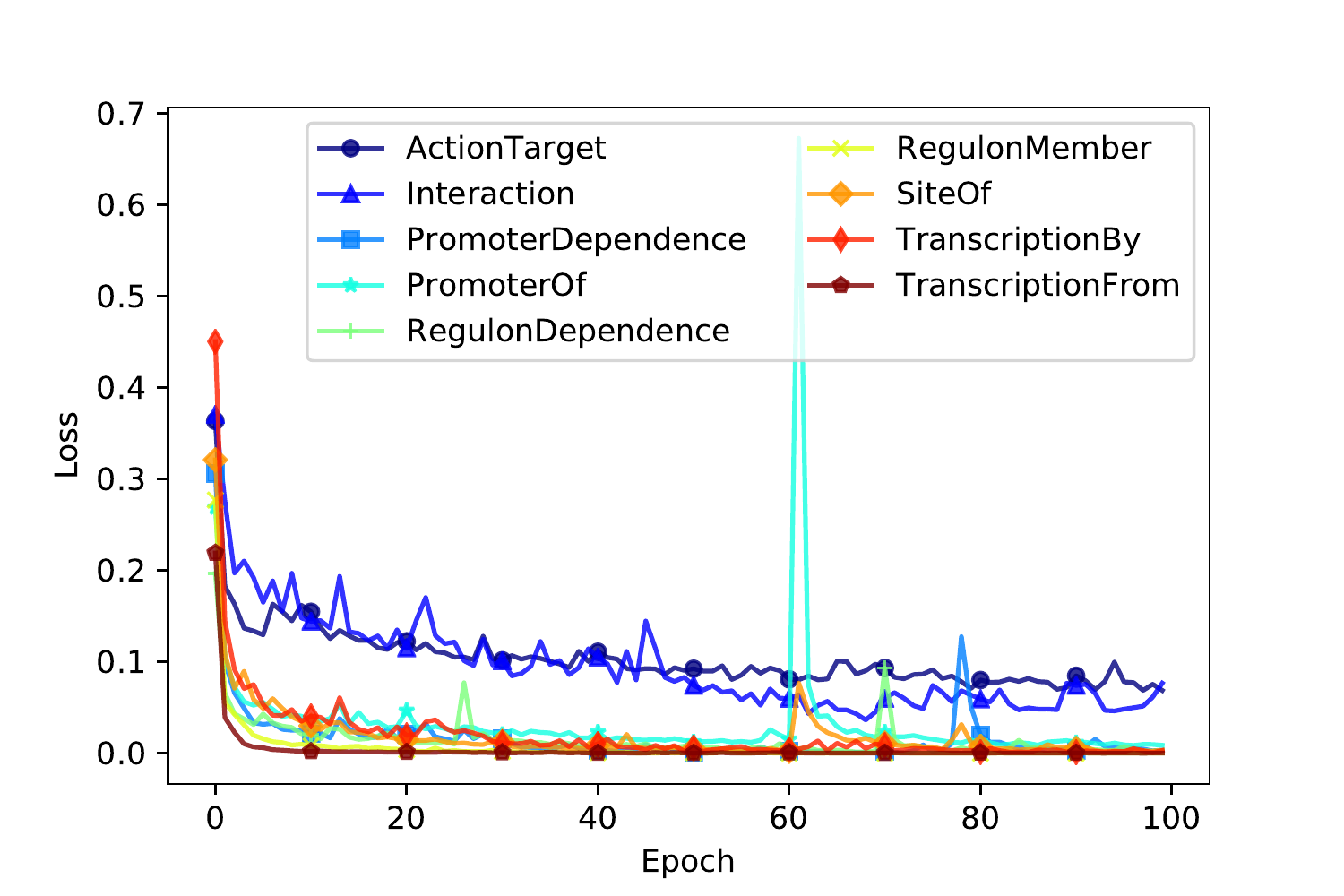}}
    \subfigure[]{\includegraphics[width=.33\textwidth]{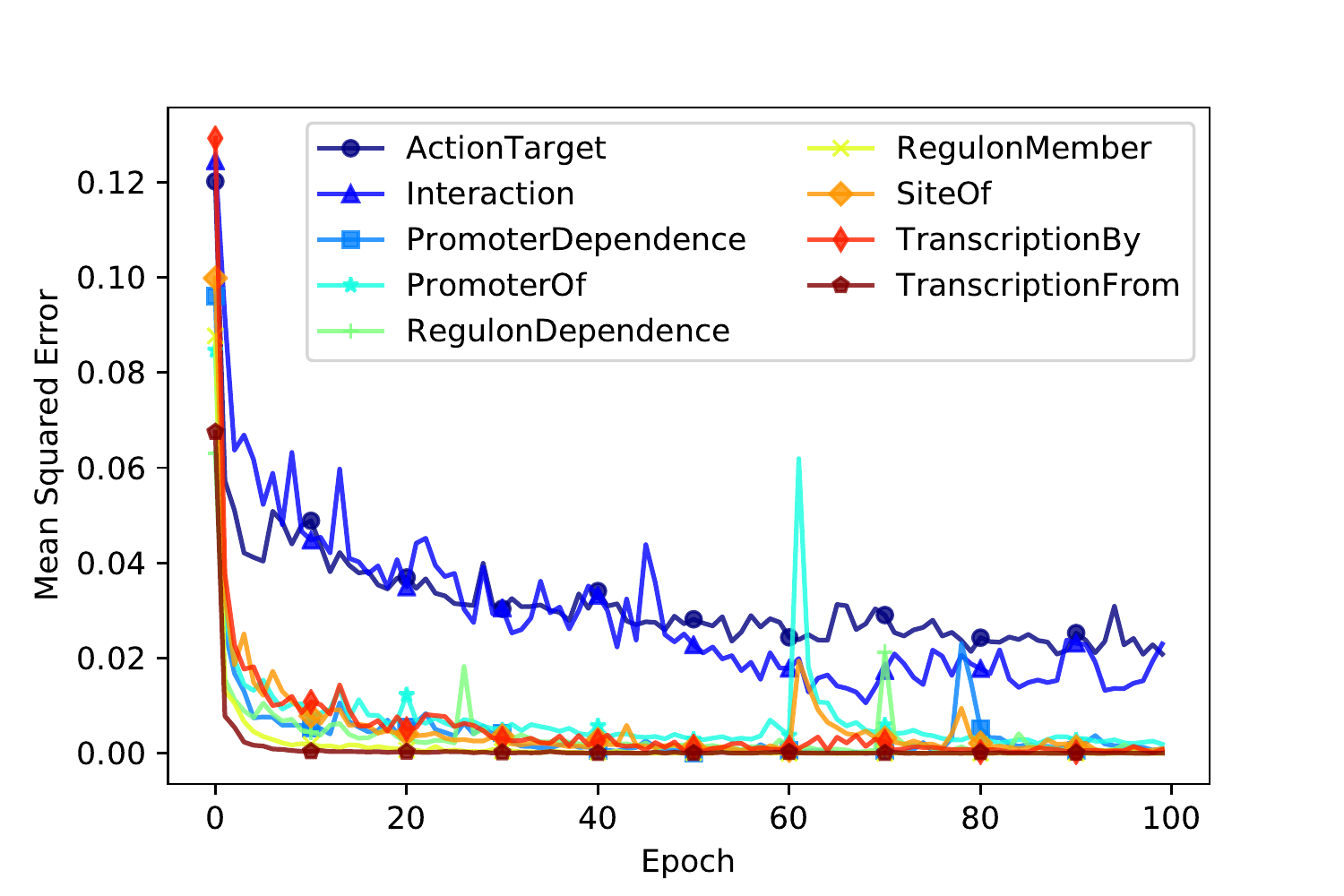}}
  }
  \caption{Performance of VeComNet on BioNLPST-BGI dataset over training and testing process. (a) Model Accuracy. (b) Model Loss. (c) Mean Squared Error.} \label{fig-bgi2011-evnt-model-perf}
\end{figure*}

\begin{figure*}[!tpb]
  \resizebox{\textwidth}{!}{\centering
    \subfigure[]{\includegraphics[width=.33\textwidth]{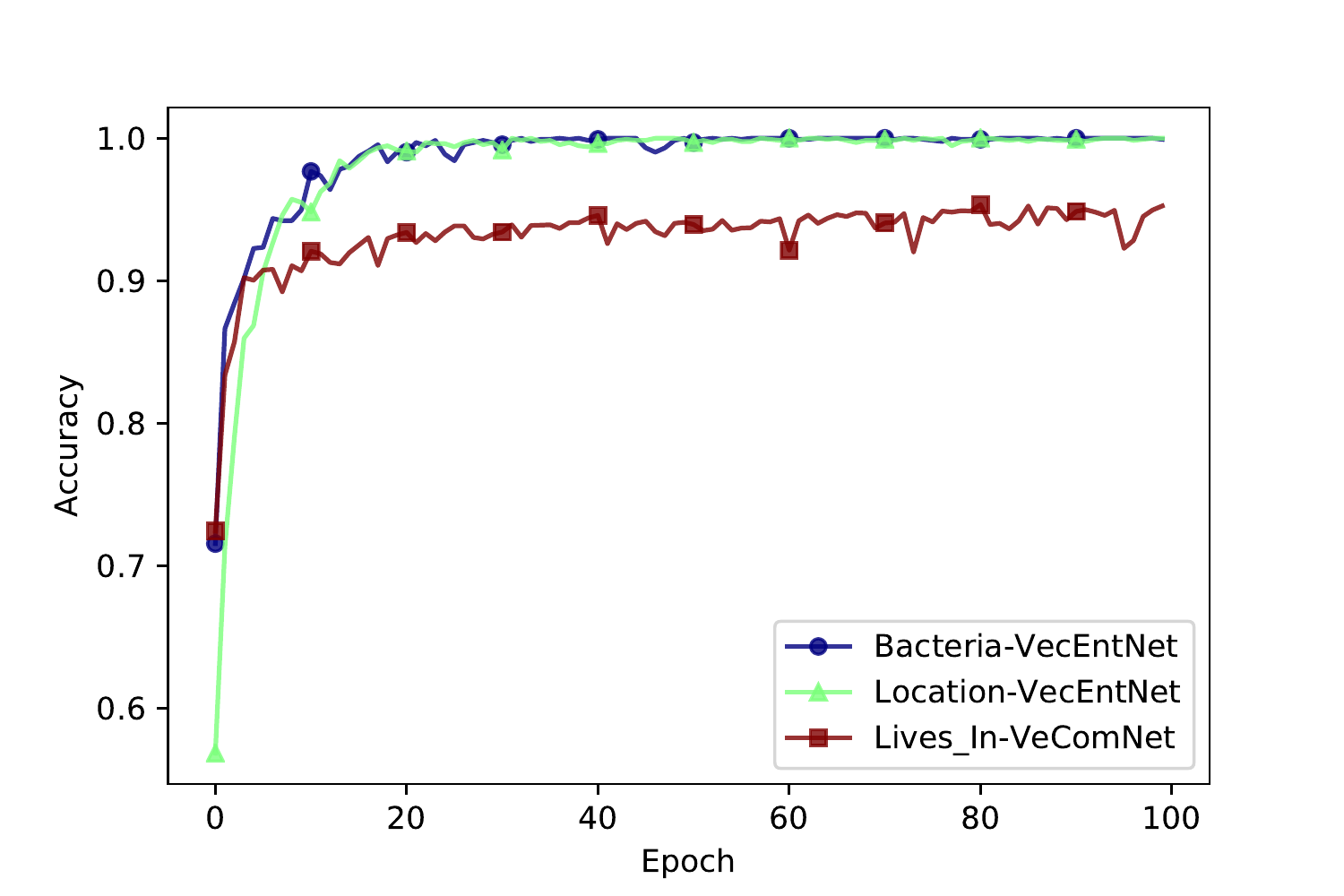}}
    \subfigure[]{\includegraphics[width=.33\textwidth]{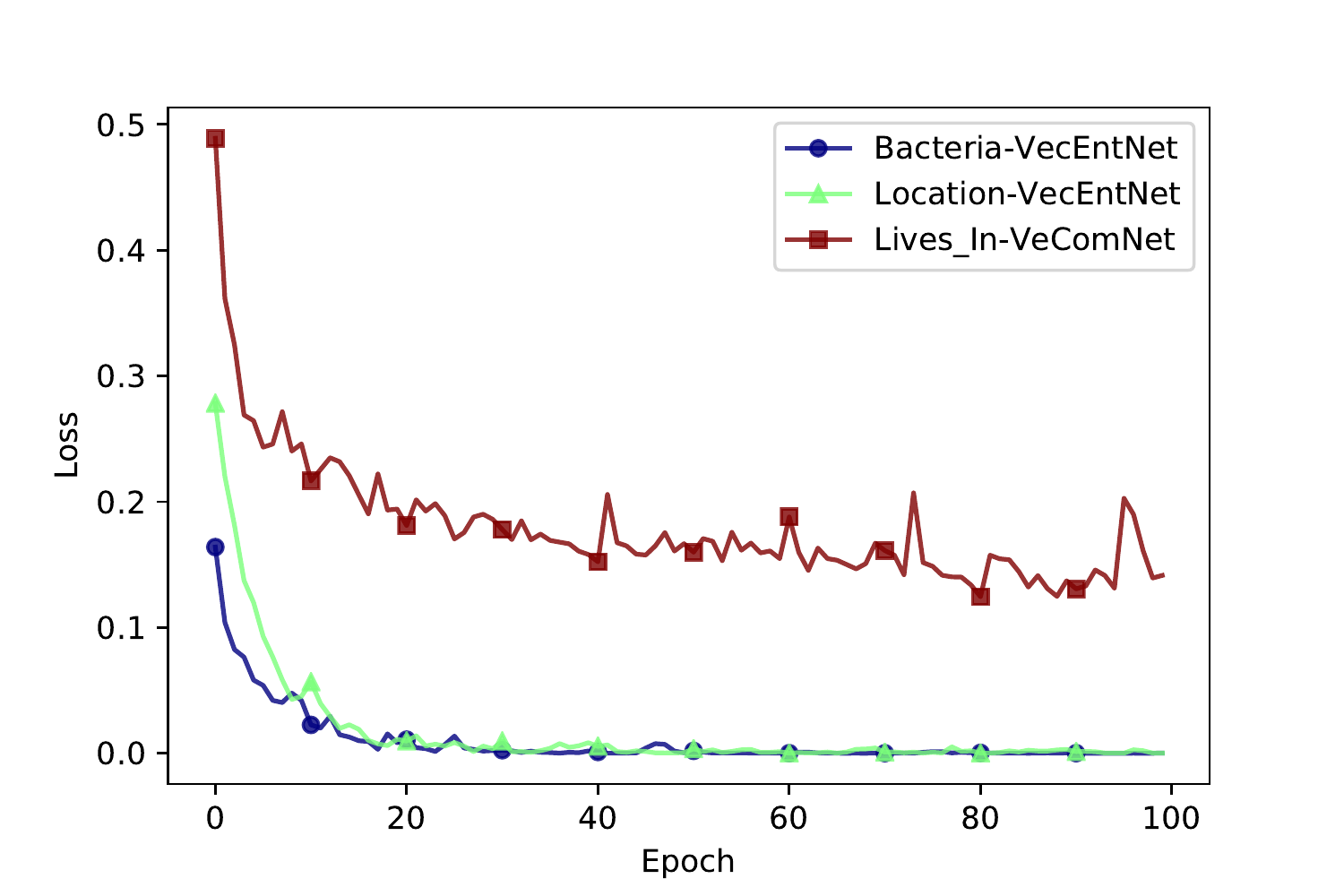}}
    \subfigure[]{\includegraphics[width=.33\textwidth]{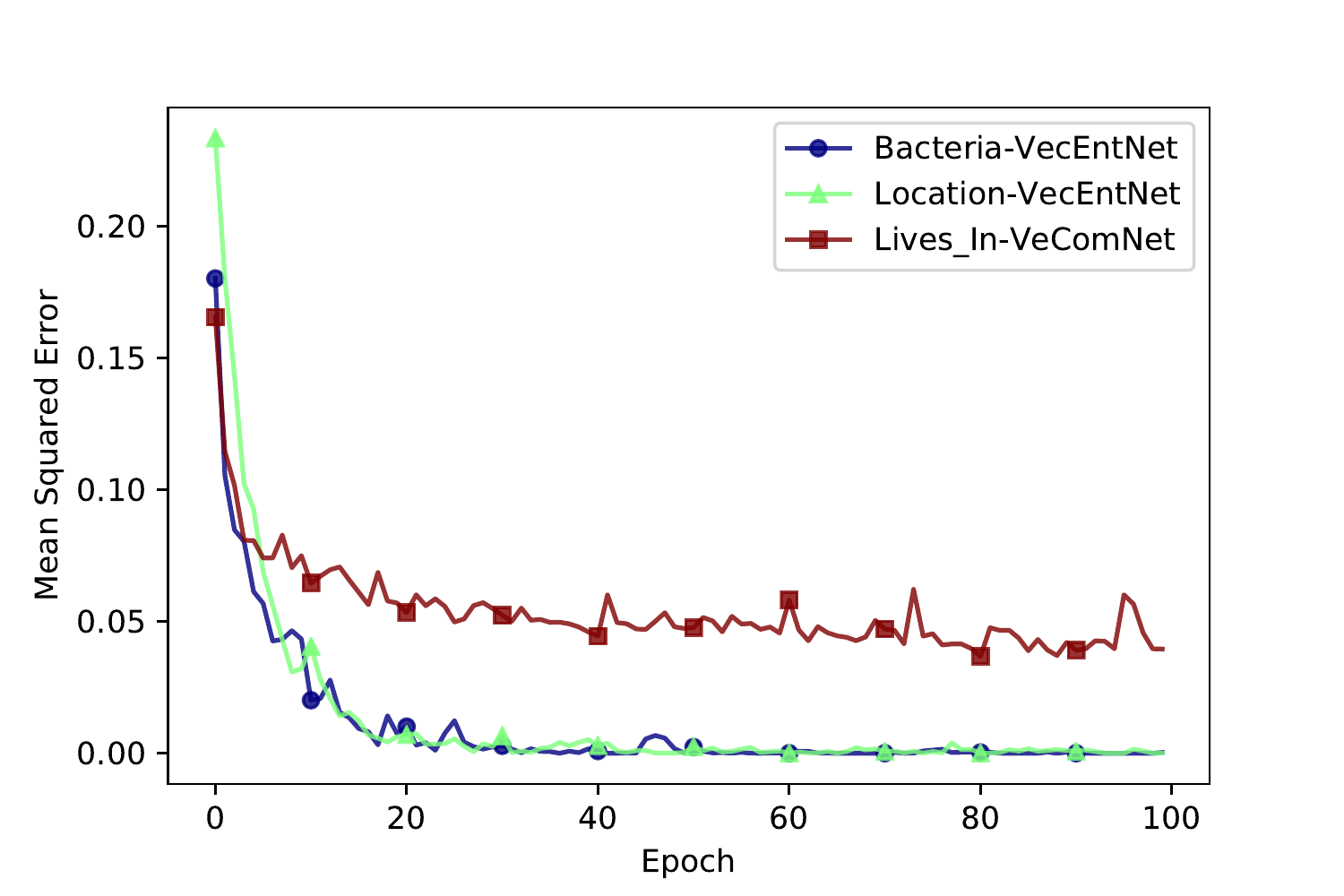}}
  }
  \caption{Performance of VecEntNet and VeComNet on BioNLPST-BB dataset over training and testing process. (a) Model Accuracy. (b) Model Loss. (c) Mean Squared Error.} \label{fig-bb2016-ent-model-perf}
\end{figure*}

We use accuracy and mean squared error to keep track of iterative training. As depicted in Figures~\ref{fig-bgi2011-ent-model-perf} and ~\ref{fig-bb2016-ent-model-perf}, VecEntNet converges roughly at the \nth{10} epoch and keeps stable in the following training. Therefore, we use 10 epochs as the default hyper-parameter in the subsequent experiments. Figure~\ref{fig-bgi2011-ent-model-perf} shows that only the argument ``Gene'' converges slower than others. 
Nevertheless, the overall performance of training VecEntNet and VeComNet is desirable.

\subsection{Performance of VecEntNet and VeComNet under 10-fold cross-validation}

\begin{figure*}[!tpb]
  \resizebox{\textwidth}{!}{\centering
    \subfigure[]{\includegraphics[width=.5\textwidth]{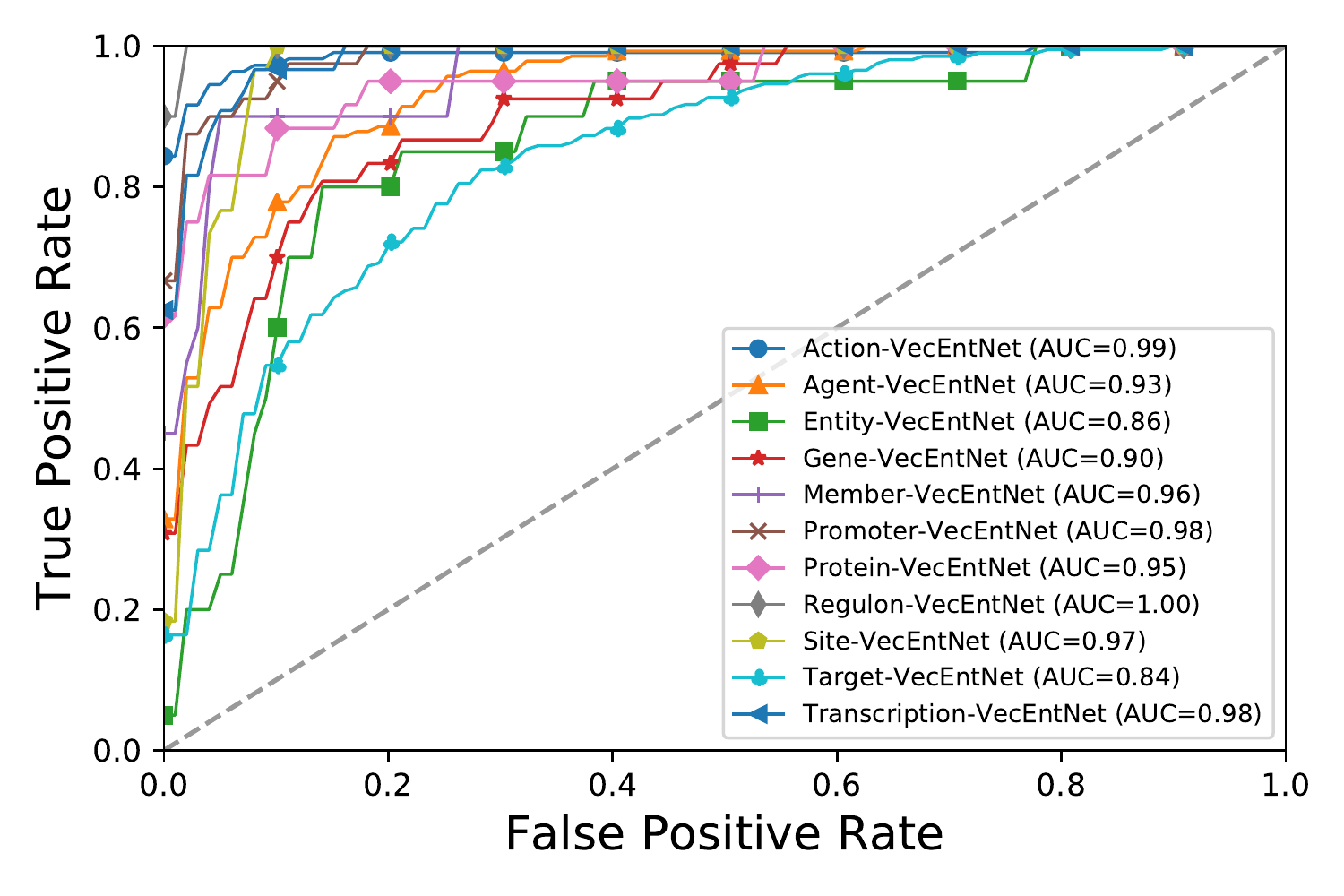}}
    \subfigure[]{\includegraphics[width=.5\textwidth]{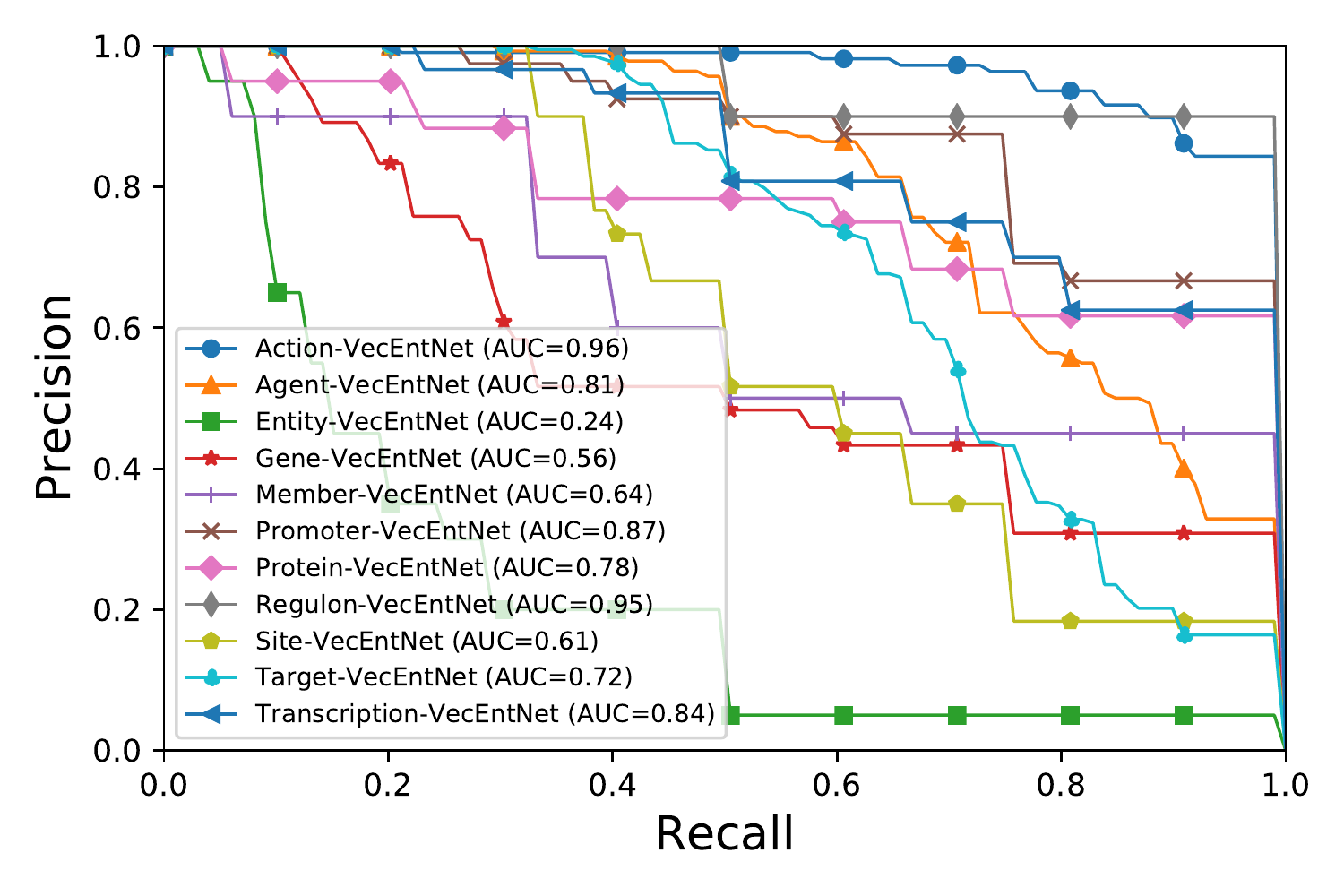}}
  }
  \caption{Performance of VecEntNet on BioNLPST-BGI dataset. (a) Micro-average ROC curves. (b) Micro-average PRC curves.} \label{fig-bgi2011-vecent-roc-prc}
\end{figure*}

\begin{figure*}[!tpb]
  \resizebox{\textwidth}{!}{\centering
    \subfigure[]{\includegraphics[width=.5\textwidth]{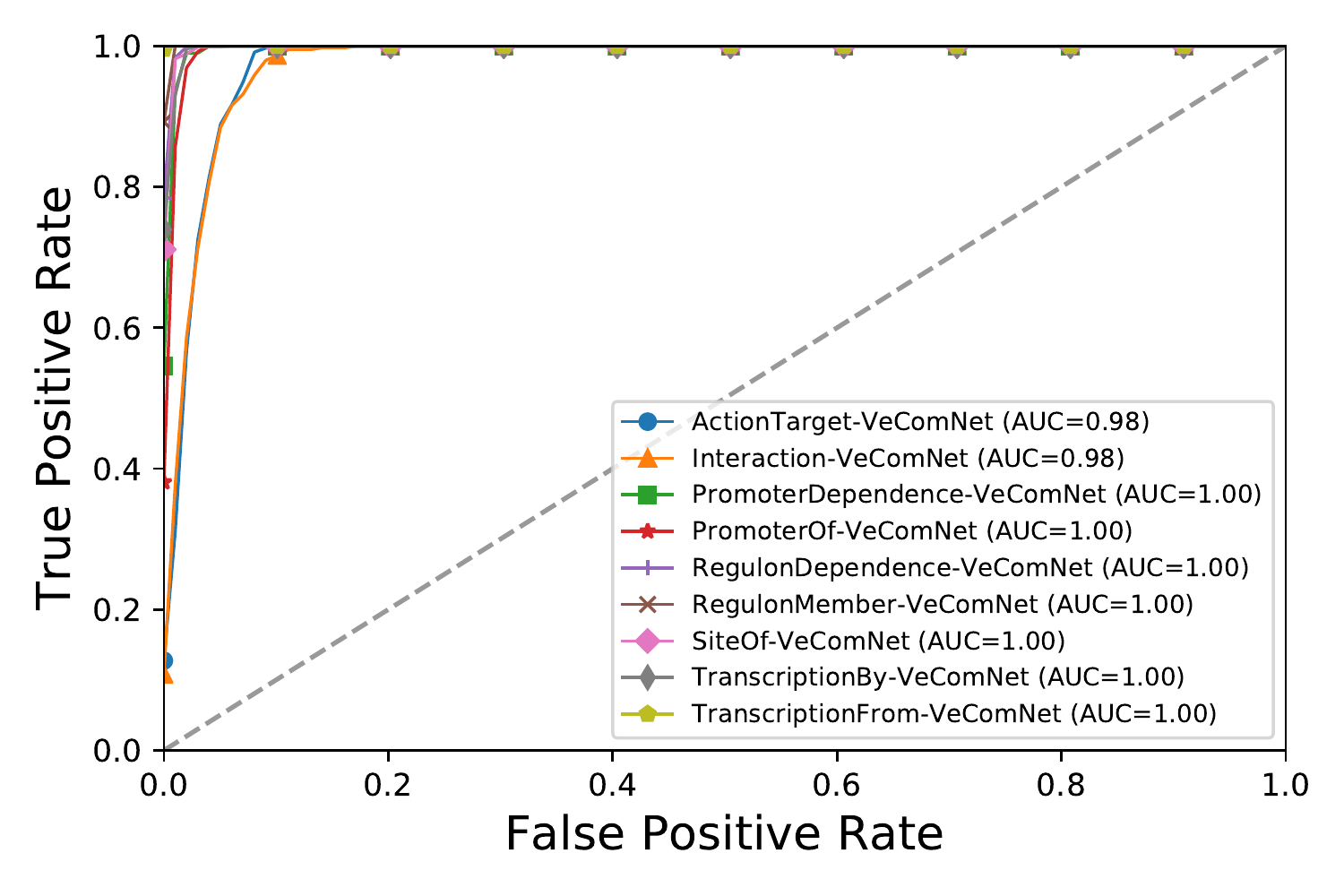}}
    \subfigure[]{\includegraphics[width=.5\textwidth]{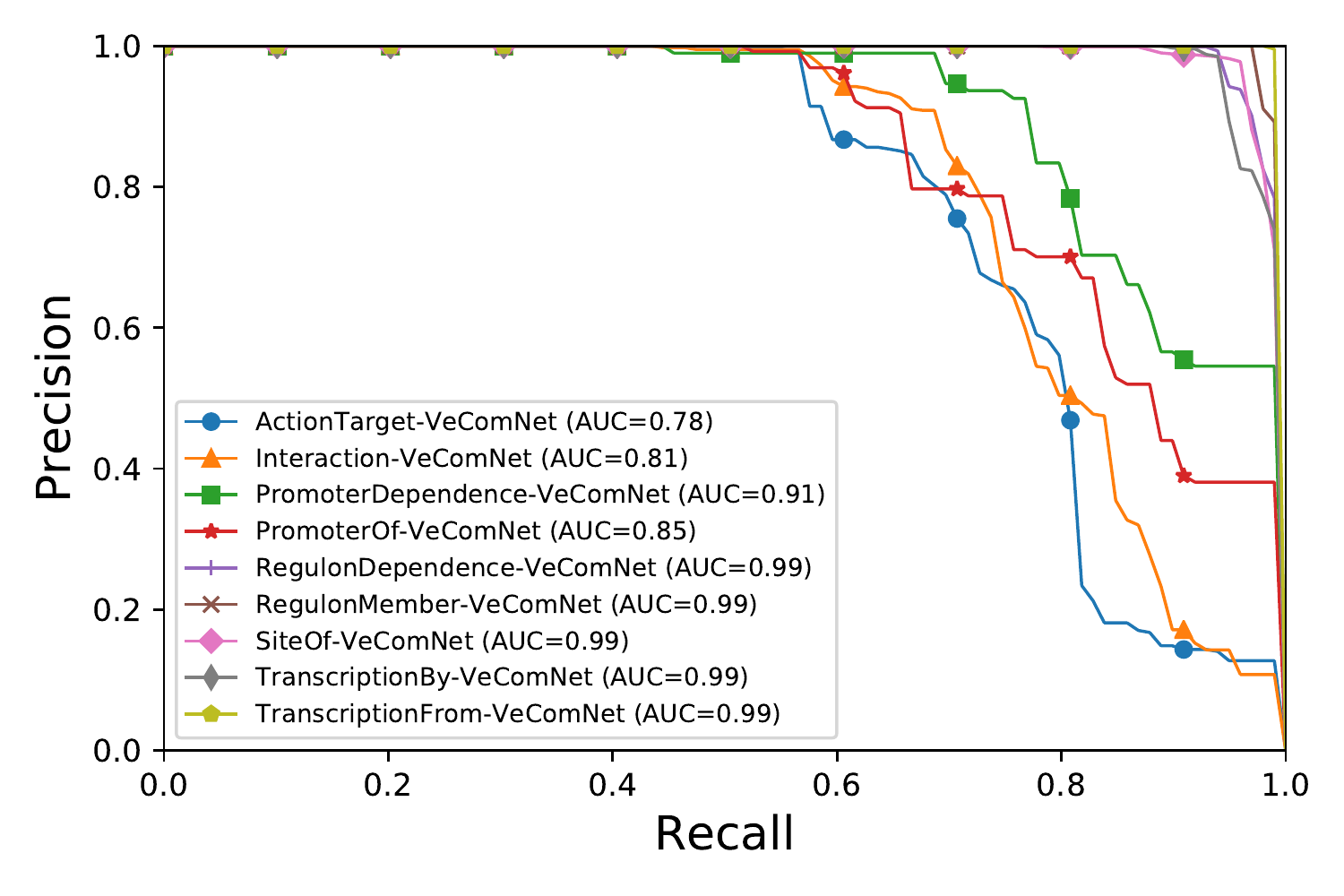}}
  }
  \caption{Performance of VeComNet on BioNLPST-BGI dataset. (a) Micro-average ROC curves. (b) Micro-average PRC curves.} \label{fig-bgi2011-vecom-roc-prc}
\end{figure*}

\begin{figure*}[!tpb]
  \resizebox{\textwidth}{!}{\centering
    \subfigure[]{\includegraphics[width=.5\textwidth]{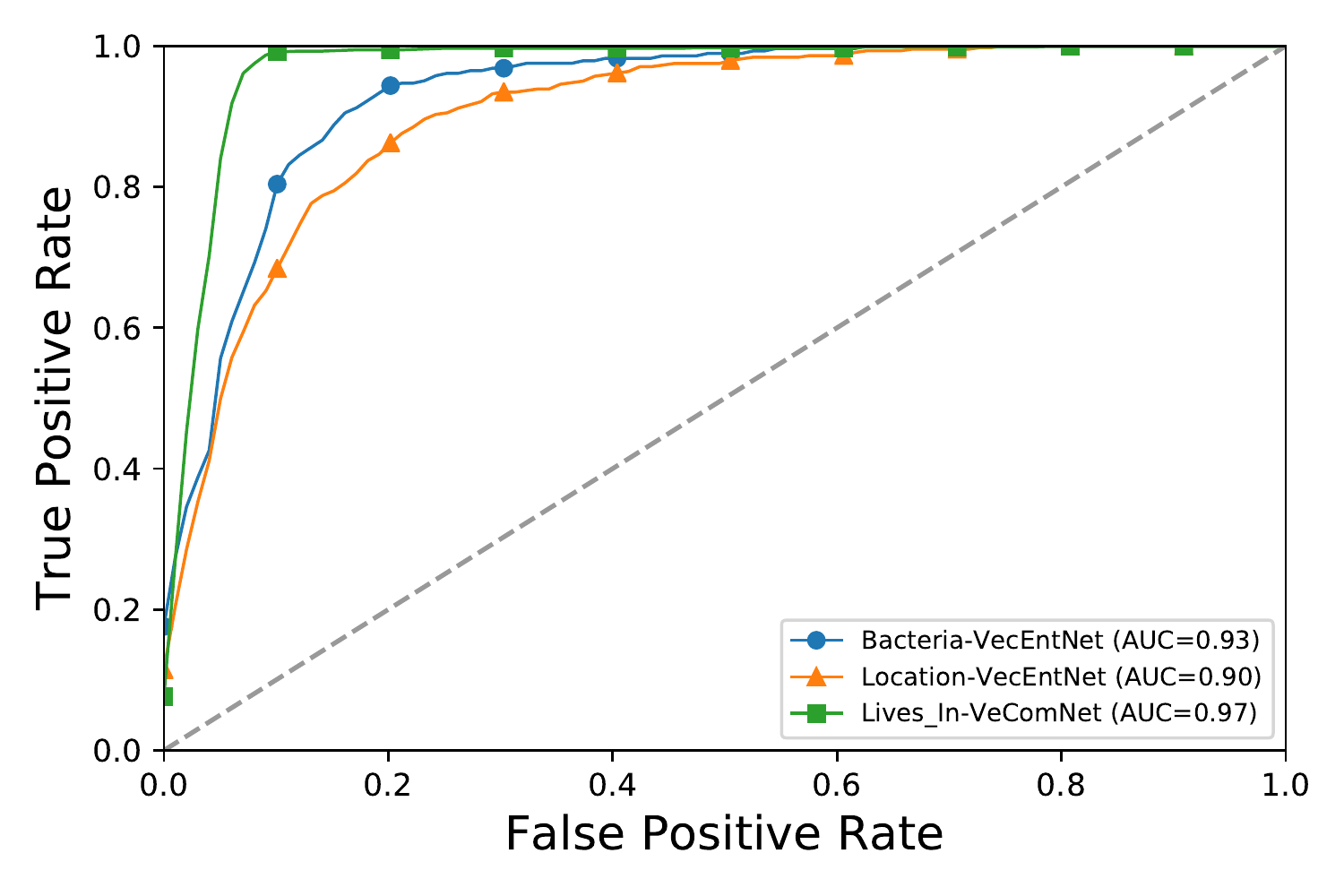}}
    \subfigure[]{\includegraphics[width=.5\textwidth]{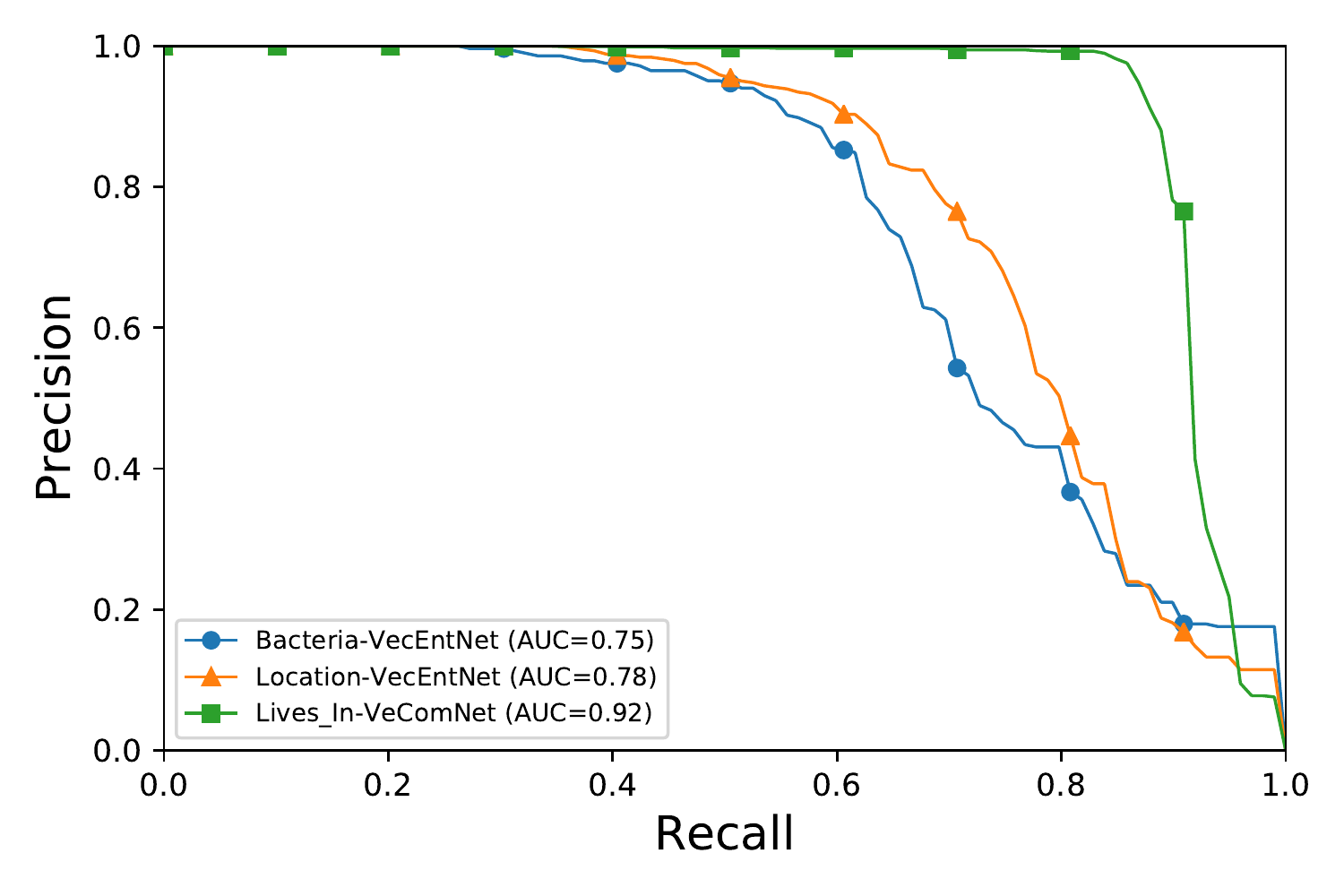}}
  }
  \caption{Performance of VecEntNet and VeComNet on BioNLPST-BB dataset. (a) Micro-average ROC curves. (b) Micro-average PRC curves.} \label{fig-bb2016-roc-prc}
\end{figure*}

We evaluate the overall performance with precision, recall, and F-score under 10-fold cross-validation experiments. We can observe from Figure~\ref{fig-bgi2011-vecent-roc-prc} that VecEntNet performs very well in most of the argument classifications on BioNLPST-BGI. However, it is expected that VecEntNet can be underestimated on the tasks with limited training samples such as ``Entity'', ``Gene'', and ``Site''. Nevertheless, VeComNet achieves robust performance by leveraging the argument embedding learned by VecEntNet. As for the performance on BioNLPST-BB dataset shown in Figure~\ref{fig-bb2016-roc-prc}, we can learn that VecEntNet as well as VeComNet performs better and keeps stable once sufficient data is given. Our proposed model definitely performs well on balanced data but it is also applicable to imbalanced labels due to the weighted loss function adopted in VecEntNet. The detailed performance is tabulated in Tables~\ref{tab-bgi2011-vecent-perf}, ~\ref{tab-bgi2011-vecom-perf}, and ~\ref{tab-bb2016-perf}

\begin{table*}[!ht]
  \caption{Performance of VecEntNet on BioNLPST-BGI dataset with regards to different argument types under 10-fold cross validation \label{tab-bgi2011-vecent-perf}} \centering {
\resizebox{\textwidth}{!}{\centering
\begin{tabular}{@{}m{\dimexpr 0.15\linewidth-2\tabcolsep}ccccccccccc@{}}
\hline
               & Action & Agent  & Entity & Gene   & Member & Promoter & Protein & Regulon & Site   & Target & Transcription \\
\hline
\rehl{\# of Training Samples} & 108 & 140 & 15 & 36 & 15 & 38 & 29 & 10 & 29 & 185 & 31 \\
Accuracy  & 0.97   & 0.83   & 0.86   & 0.91   & 0.97   & 0.97     & 0.92    & 0.99    & 0.94   & 0.77   & 0.97          \\
Precision & 0.92   & 0.56   & 0.09   & 0.32   & 0.48   & 0.70     & 0.52    & 0.68    & 0.45   & 0.61   & 0.60          \\
Recall    & 0.93   & 0.93   & 0.65   & 0.55   & 0.80   & 0.90     & 0.72    & 1.00    & 0.80   & 0.73   & 0.98          \\
F score   & 0.92   & 0.70   & 0.15   & 0.37   & 0.54   & 0.77     & 0.50    & 0.77    & 0.52   & 0.66   & 0.75          \\
Train time (s) & 603.73 & 696.15 & 512.58 & 710.31 & 677.00 & 469.79   & 595.32  & 391.12  & 591.96 & 660.29 & 610.50        \\
Test time (s)  & 1.65   & 1.59   & 1.56   & 1.79   & 2.15   & 0.89     & 1.02    & 0.82    & 0.94   & 1.87   & 1.09          \\
\hline
  	\end{tabular}}}{}
\end{table*}

\begin{table*}[!ht]
  \caption{Performance of VeComNet on BioNLPST-BGI dataset with regards to different event types under 10-fold cross validation \label{tab-bgi2011-vecom-perf}} {
\resizebox{\textwidth}{!}{\centering
\begin{tabular}{@{}m{\dimexpr 0.1\linewidth-2\tabcolsep}cccccccccc@{}}
\hline
               & ActionTarget & Interaction & PromoterDependence & PromoterOf & RegulonDependence & RegulonMember & SiteOf & TranscriptionBy & TranscriptionFrom \\
\hline
Accuracy  & 0.93   & 0.91        & 0.98               & 0.98       & 0.99              & 0.99          & 0.98   & 0.97            & 0.99              \\
Precision & 0.70   & 0.73        & 0.82               & 0.79       & 0.97              & 0.99          & 0.95   & 0.60            & 0.99              \\
Recall    & 0.91   & 0.82        & 0.84               & 0.78       & 0.99              & 0.99          & 0.98   & 0.98            & 0.99              \\
F score   & 0.79   & 0.77        & 0.82               & 0.76       & 0.98              & 0.99          & 0.97   & 0.75            & 0.99              \\
Train time (s) & 4.99   & 5.04        & 5.00               & 5.03       & 5.55              & 5.67          & 6.18   & 5.69            & 5.78              \\
Test time (s)  & 0.15   & 0.15        & 0.16               & 0.16       & 0.16              & 0.17          & 0.16   & 0.16            & 0.14              \\
\hline
  	\end{tabular}}}{}
\end{table*}

\begin{table}[!ht]
  \caption{Performance of VecEntNet and VeComNet on BioNLPST-BB dataset \label{tab-bb2016-perf}} \centering {
\begin{tabular}{@{}m{\dimexpr 0.3\linewidth-2\tabcolsep}ccc@{}}
\hline
\multirow{2}{*}{} & \multicolumn{2}{c}{VecEntNet} & VeComNet  \\ \cline{2-4} 
                  & Bacteria      & Location      & Lives\_In \\
\hline
Accuracy     & 0.88          & 0.82          & 0.92      \\
Precision    & 0.66          & 0.69          & 0.89      \\
Recall       & 0.74          & 0.77          & 0.96      \\
F score      & 0.69          & 0.72          & 0.92      \\
Train time (s)    & 771.44        & 757.76        & 4.83      \\
Test time (s)     & 0.72          & 0.74          & 0.15      \\
\hline
  	\end{tabular}}{}
\end{table}

As for the two worst cases of argument classification, ``Entity'' and ``Gene'' (F scores = 0.15 and 0.37), their corresponding event detection is still satisfactory (F scores = 0.97 and 0.76). We can also observe that the argument with better performance (``Site'' and ``Promoter'') within the same event type compensate the defectiveness of the worse one.

\subsection{Performance comparison with other top-ranked approaches}

\begin{table}[!ht]
  \caption{Performance comparison between VeComNet and the \textbf{best} method (Uturku) on BioNLPST-BGI dataset \label{tab-bgi2011-perf-cmp}} \centering {
\begin{tabular}{lcccccc}
\hline
Method & \multicolumn{3}{c}{VeComNet} & \multicolumn{3}{c}{Uturku \cite{Bjorne2012}} \\
\hline
Event Type & Precision & Recall & F-score & Precision & Recall & F-score \\
\hline
ActionTarget & 0.7 & 0.91 & 0.79 & 0.94 & 0.92 & 0.93 \\
Interaction & 0.73 & \textbf{0.82} & \textbf{0.77} & 0.75 & 0.56 & 0.64 \\
PromoterDependence & 0.82 & 0.84 & 0.82 & 1.00 & 1.00 & 1.00 \\
PromoterOf & 0.79 & 0.78 & 0.76 & 1.00 & 1.00 & 1.00 \\
RegulonDependence & 0.97 & 0.99 & 0.98 & 1.00 & 1.00 & 1.00 \\
RegulonMember & 0.99 & \textbf{0.99} & \textbf{0.99} & 1.00 & 0.50 & 0.67 \\
SiteOf & 0.95 & \textbf{0.98} & \textbf{0.97} & 1.00 & 0.17 & 0.29 \\
TranscriptionBy & \textbf{0.60} & \textbf{0.98} & \textbf{0.75} & 0.67 & 0.50 & 0.57 \\
TranscriptionFrom & 0.99 & 0.99 & 0.99 & 1.00 & 1.00 & 1.00 \\
Average & 0.788 & \textbf{0.834} & \textbf{0.791} & 0.91 & 0.83 & 0.79 \\
\hline
  	\end{tabular}}{}
\end{table}

We compared our performance with that of the best method in the competition on BioNLPST-BGI dataset with respect to each event type. As tabulated in Table~\ref{tab-bgi2011-perf-cmp}, VeComNet and the Uturku's approach \cite{Bjorne2014,Bjorne2015} have their own merits. \rehl{VeComNet performs the best on ``Interaction'', ``RegulonMember'', ``SiteOf'', ``TranscriptionBy'' events with significant improvement on the F-scores (0.12, 0.32, 0.68, 0.4) compared to the best existing approach; and has competitive performance on ``RegulonDependence'' and ``TranscriptionFrom'' events.} 
The performance of VeComNet on other events are stable and impressive due to which its average performance is better than the Uturku's approach. \rehl{The compared method from Uturku seems over-fitting to the dataset since in most of the event types it achieved the ideal F-scores of 1.0 where our proposed method does not. Our method outstands from other approaches according to its generalization ability instead of the ideal F-scores.} But the deep learning model adopted in VeComNet is limited by the number of training samples.

\begin{table}[!ht]
  \caption{Performance comparison between VeComNet and other top-ranked methods \cite{Deleger2016} on BioNLPST-BB dataset \label{tab-bb2016-perf-cmp}} \centering {
\begin{tabular}{lccc}
\hline
Method & Precision & Recall & F-score \\
\hline
VeComNet & \textbf{0.89} & \textbf{0.96} & \textbf{0.92} \\
VERSE \cite{Lever2016} & 0.51 & 0.62 & 0.56 \\
TurkuNLP \cite{Mehryary2016} & 0.63 & 0.45 & 0.52 \\
LIMSI & 0.39 & 0.65 & 0.49 \\
HK & 0.60 & 0.39 & 0.47 \\
whunlpre & 0.56 & 0.41 & 0.47 \\
DUTIR \cite{Li2016} & 0.57 & 0.38 & 0.46 \\
WXU & 0.56 & 0.38 & 0.46 \\
\hline
  	\end{tabular}}{}
\end{table}

From Table~\ref{tab-bb2016-perf-cmp} we can observe that VeComNet has the strongest power in single event prediction. The less arguments and event types contained in the detection task, the more powerful VeComNet will be. Besides, VeComNet is a generic model that can be used in different event detection tasks without any tuning and modification. The robustness and strong predictive power of VeComNet enables it to be a promising model in the area of biomedical event extraction.

\section{Case studies}

To reveal how our method works, we randomly picked some cases from the testing dataset for case demonstration. The sample sentence `The expression of rsfA is under the control of both sigma(F) and sigma(G).' with ID `PMID-10629188-S5' in the testing dataset of BioNLPST-BGI has four recognized entities ($T_{1}$:`expression', $T_{2}$:`rsfA', $T_{3}$:`sigma(F)', $T_{4}$:`sigma(G)') and three events (ActionTarget: $[Action]T_{1}\longrightarrow[Target]T_{2}$, Interaction: $[Agent]T_{3}\longrightarrow[Target]T_{2}$, Interaction: $[Agent]T_{4}\longrightarrow[Target]T_{2}$) as ground true annotations. We obtained 11 argument models by fitting VecEntNet on the training dataset with the entity annotations. We further gained the argument embeddings for each possible pair of entities in both training and testing datasets. For the above-mentioned sample, some of the candidate pairs generated are $<T_{1}, T_{2}>, <T_{2}, T_{1}>, <T_{2}, T_{3}>, <T_{3}, T_{2}>, <T_{2}, T_{4}>, <T_{4}, T_{2}>, <T_{1}, T_{3}>, <T_{1}, T_{4}>$. And the argument models for event type ActionTarget are $arg_{action}$ and $arg_{target}$. \rehl{We take them as functions and the candidate pairs of entities as input.} The argument embeddings we obtained for $<T_{1}, T_{2}>$ are $<arg_{action}(T_{1}) \oplus arg_{target}(T_{1}), arg_{target}(T_{2}) \oplus arg_{action}(T_{2})>$. Since we are not aware of the argument type the entities belong to, we concatenated both argument embeddings for each entity and let VeComNet to determine. The argument embeddings are obtained for other candidate entity pairs with respect to different event types in a similar way. We used argument embeddings as the input of VeComNet models. The predicted labels for the aforementioned candidate entity pairs are $<1,1>,<1,0>,<0,0>...<0,0>$ with respect to ActionTarget event and $<0,0>,<0,0>,<1,0>,<1,1>,<1,0>,<1,1>,<0,0>,<0,0>$ with respect to Interaction event, in which the first label indicates the existence of the corresponding event and the second label indicates whether the event is pointed from the first entity to the second one. The binary labels were further post-processed to generate the predicted biomedical events. \rehl{For instance, the candidate pairs $<T_{1}, T_{2}>$ and $<T_{2}, T_{3}>$ are predicted as $<1,1>$ and $<1,0>$ for ActionTarget and Interaction evnets respectively. It means that it exists an ActionTarget event $T_{1} \longrightarrow T_{2}$ (expression-\textgreater{}rsfA) and an Interaction event $T_{3} \longrightarrow T_{2}$ (rsfA-\textgreater{}sigma(F)) in this sentence.}

\section{Discussion}

For many years, scientific literature has served as the major outlet for novel discovery and result dissemination. To extract useful knowledge from the literature for management and query, information extraction is proposed to automate this process. Biomedical event extraction is particularly important because it is able to systematically organize the knowledge as controlled representations such as directed knowledge graphs. However, the existing event detection methods are not satisfactory in performance because most of them are constrained in the trigger-based approach which relies on the lexical and syntactic features from dependency parsing. The quality of manual trigger annotation and the error propagation from trigger detection to the event detection have limited our progress for years.

In this study, we proposed a bottom-up event detection framework using deep learning techniques. We built an LSTM-based model VecEntNet to construct argument embeddings for each recognized entity. We further utilized compositional attributes of the argument vectors to train a directed event classifier VeComNet.

\rehl{LSTM and context embedding have been shown its applicability in several other NLP tasks. Our main contribution is the proposed framework for argument embedding using Bi-directional LSTM and the downstream directed event detection using multi-output neural network. This strategy for event detection is proposed for the first time in this study. It overcomes the error propagation as well as extra annotations of trigger-based approaches. Besides, the continuous space of argument embedding significantly lessen the sensitivity of event detection. In addition, we developed our own loss functions for training the argument embedding with unbalanced data and training the multi-output neural network for directed event detection. These are the key reasons why our method can achieve outstanding performance. Broadly speaking, the proposed method is suitable for general event extraction by using the pre-trained word embedding in the specific area.}

Our method is not sensitive to the hyper-parameters and it works well for a wide range of tasks. The experimental results indicate that the proposed method is competent in the biomedical event extraction. In the future, we envision that it can fundamentally benefit the related downstream tasks in biomedical text mining with broad impacts. 

\section*{Acknowledgements}

The authors 
are grateful to the organizers of BioNLP Shared Task who provide the public annotated dataset. The authors would also like to thank Prashant Sridhar for his English proofreading. \vspace*{-12pt}

\section*{Funding}
The work described in this paper was substantially supported by three grants from the Research Grants Council of the Hong Kong Special Administrative Region [CityU 21200816], [CityU 11203217], and [CityU 11200218]. We acknowledge the donation support of the Titan Xp GPU from the NVIDIA Corporation. \vspace*{-12pt}

\bibliographystyle{unsrt}



\begin{thebibliography}{10}

\bibitem{Rebholz-Schuhmann2012}
Dietrich Rebholz-Schuhmann, Anika Oellrich, and Robert Hoehndorf.
\newblock {Text-mining solutions for biomedical research: enabling integrative
  biology}.
\newblock {\em Nature Reviews Genetics}, 13(12):829--839, dec 2012.

\bibitem{Mallory2015}
Emily~K Mallory, Ce~Zhang, Christopher R{\'{e}}, and Russ~B Altman.
\newblock {Large-scale extraction of gene interactions from full text
  literature using DeepDive}.
\newblock {\em Bioinformatics}, 2015.

\bibitem{Zhao2016}
Zhehuan Zhao, Zhihao Yang, Ling Luo, Hongfei Lin, and Jian Wang.
\newblock {Drug drug interaction extraction from biomedical literature using
  syntax convolutional neural network}.
\newblock {\em Bioinformatics}, 32(22):btw486, jul 2016.

\bibitem{Perfetto2015}
Livia Perfetto, Leonardo Briganti, Alberto Calderone, Andrea {Cerquone
  Perpetuini}, Marta Iannuccelli, Francesca Langone, Luana Licata, Milica
  Marinkovic, Anna Mattioni, Theodora Pavlidou, Daniele Peluso, Lucia~Lisa
  Petrilli, Stefano Pirr{\`{o}}, Daniela Posca, Elena Santonico, Alessandra
  Silvestri, Filomena Spada, Luisa Castagnoli, Gianni Cesareni, Andrea~Cerquone
  Perpetuini, Marta Iannuccelli, Francesca Langone, Luana Licata, Milica
  Marinkovic, Anna Mattioni, Theodora Pavlidou, Daniele Peluso, Lucia~Lisa
  Petrilli, Stefano Pirr{\`{o}}, Daniela Posca, Elena Santonico, Alessandra
  Silvestri, Filomena Spada, Luisa Castagnoli, and Gianni Cesareni.
\newblock {SIGNOR: a database of causal relationships between biological
  entities}.
\newblock {\em Nucleic Acids Research}, 44(D1):D548--D554, jan 2015.

\bibitem{Lim2016}
Kun Ming~Kenneth Lim, Chenhao Li, Kern~Rei Chng, and Niranjan Nagarajan.
\newblock {@MInter: automated text-mining of microbial interactions}.
\newblock {\em Bioinformatics}, 32(19):2981--2987, oct 2016.

\bibitem{Canada2017}
Andres Ca{\~{n}}ada, Salvador Capella-Gutierrez, Obdulia Rabal, Julen
  Oyarzabal, Alfonso Valencia, and Martin Krallinger.
\newblock {LimTox: a web tool for applied text mining of adverse event and
  toxicity associations of compounds, drugs and genes}.
\newblock {\em Nucleic Acids Research}, 45(W1):W484--W489, jul 2017.

\bibitem{Bui2012}
Q.-C. Bui and P.~M.~A. Sloot.
\newblock {A robust approach to extract biomedical events from literature}.
\newblock {\em Bioinformatics}, 28(20):2654--2661, oct 2012.

\bibitem{Bjorne2010}
J.~Bjorne, F.~Ginter, S.~Pyysalo, J.~Tsujii, and T.~Salakoski.
\newblock {Complex event extraction at PubMed scale}.
\newblock {\em Bioinformatics}, 26(12):i382--i390, jun 2010.

\bibitem{Bjorne2011}
Jari Bj{\"{o}}rne and Tapio Salakoski.
\newblock {Generalizing biomedical event extraction}.
\newblock {\em Proceedings of the BioNLP Shared Task 2011 Workshop}, pages
  183--191, 2011.

\bibitem{Ananiadou2010}
Sophia Ananiadou, Sampo Pyysalo, Jun'ichi Tsujii, and Douglas~B. Kell.
\newblock {Event extraction for systems biology by text mining the literature}.
\newblock {\em Trends in Biotechnology}, 28(7):381--390, jul 2010.

\bibitem{Pyysalo2012}
Sampo Pyysalo, Tomoko Ohta, Makoto Miwa, Han-Cheol Cho, Jun'ichi Tsujii, and
  Sophia Ananiadou.
\newblock {Event extraction across multiple levels of biological organization}.
\newblock {\em Bioinformatics}, 28(18):i575--i581, sep 2012.

\bibitem{Nguyen2016}
Thien~Huu Nguyen, Kyunghyun Cho, and Ralph Grishman.
\newblock {Joint event extraction via recurrent neural networks}.
\newblock In {\em Proceedings of the 2016 Conference of the North American
  Chapter of the Association for Computational Linguistics: Human Language
  Technologies}, pages 300--309, 2016.

\bibitem{Mikolov2013}
Tomas Mikolov, Ilya Sutskever, Kai Chen, Greg~S Corrado, and Jeff Dean.
\newblock {Distributed Representations of Words and Phrases and their
  Compositionality}.
\newblock In C~J~C Burges, L~Bottou, M~Welling, Z~Ghahramani, and K~Q
  Weinberger, editors, {\em Advances in Neural Information Processing Systems
  26}, pages 3111--3119. Curran Associates, Inc., 2013.

\bibitem{Zhou2014}
Deyu Zhou, Dayou Zhong, and Yulan He.
\newblock {Event trigger identification for biomedical events extraction using
  domain knowledge}.
\newblock {\em Bioinformatics}, 30(11):1587--1594, jun 2014.

\bibitem{Kim2009}
Jin-Dong Kim, Tomoko Ohta, Sampo Pyysalo, Yoshinobu Kano, and Jun'ichi Tsujii.
\newblock {Overview of BioNLP'09 shared task on event extraction}.
\newblock In {\em Proceedings of the Workshop on Current Trends in Biomedical
  Natural Language Processing: Shared Task}, pages 1--9. Association for
  Computational Linguistics, 2009.

\bibitem{Kim2011}
Jin-Dong Kim, Sampo Pyysalo, Tomoko Ohta, Robert Bossy, Ngan Nguyen, and
  Jun'ichi Tsujii.
\newblock {Overview of BioNLP shared task 2011}.
\newblock In {\em Proceedings of the BioNLP shared task 2011 workshop}, pages
  1--6. Association for Computational Linguistics, 2011.

\bibitem{Nedellec2013}
Claire N{\'{e}}dellec, Robert Bossy, Jin-Dong Kim, Jung-jae Kim, Tomoko Ohta,
  Sampo Pyysalo, and Pierre Zweigenbaum.
\newblock {Overview of BioNLP Shared Task 2013}, 2013.

\bibitem{Nedellec2016}
Claire N{\'{e}}dellec, Robert Bossy, and Jin-Dong Kim.
\newblock {Proceedings of the 4th BioNLP Shared Task Workshop}.
\newblock In {\em Proceedings of the 4th BioNLP Shared Task Workshop}, 2016.

\bibitem{Nivre2016}
Joakim Nivre, Marie-Catherine de~Marneffe, Filip Ginter, Yoav Goldberg, Jan
  Hajic, Christopher~D Manning, Ryan~T McDonald, Slav Petrov, Sampo Pyysalo,
  and Natalia Silveira.
\newblock {Universal Dependencies v1: A Multilingual Treebank Collection.}
\newblock In {\em LREC}, 2016.

\bibitem{Chen2014}
Danqi Chen and Christopher Manning.
\newblock {A fast and accurate dependency parser using neural networks}.
\newblock In {\em Proceedings of the 2014 conference on empirical methods in
  natural language processing (EMNLP)}, pages 740--750, 2014.

\bibitem{McClosky2008}
David McClosky and Eugene Charniak.
\newblock {Self-training for biomedical parsing}.
\newblock In {\em Proceedings of the 46th Annual Meeting of the Association for
  Computational Linguistics on Human Language Technologies: Short Papers},
  pages 101--104. Association for Computational Linguistics, 2008.

\bibitem{Luo2017}
Yuan Luo, {\"{O}}zlem Uzuner, and Peter Szolovits.
\newblock {Bridging semantics and syntax with graph
  algorithms—state-of-the-art of extracting biomedical relations}.
\newblock {\em Briefings in Bioinformatics}, 18(1):160--178, jan 2017.

\bibitem{Jagannatha2016}
Abhyuday~N Jagannatha and Hong Yu.
\newblock {Bidirectional RNN for Medical Event Detection in Electronic Health
  Records.}
\newblock {\em Proceedings of the conference. Association for Computational
  Linguistics. North American Chapter. Meeting}, 2016:473--482, jun 2016.

\bibitem{Bengio2003}
Yoshua Bengio, R{\'{e}}jean Ducharme, Pascal Vincent, and Christian Jauvin.
\newblock {A neural probabilistic language model}.
\newblock {\em Journal of machine learning research}, 3(Feb):1137--1155, 2003.

\bibitem{Melamud2016}
Oren Melamud, Jacob Goldberger, and Ido Dagan.
\newblock {context2vec: Learning generic context embedding with bidirectional
  lstm}.
\newblock In {\em Proceedings of The 20th SIGNLL Conference on Computational
  Natural Language Learning}, pages 51--61, 2016.

\bibitem{Kosmopoulos2015}
Aris Kosmopoulos, Ion Androutsopoulos, and Georgios Paliouras.
\newblock {Biomedical semantic indexing using dense word vectors in bioasq}.
\newblock {\em J BioMed Semant Suppl BioMedl Inf Retr},
  3410:959136040--1510456246, 2015.

\bibitem{Gers1999}
F.A. Gers.
\newblock {Learning to forget: continual prediction with LSTM}.
\newblock In {\em 9th International Conference on Artificial Neural Networks:
  ICANN '99}, volume 1999, pages 850--855. IEE, 1999.

\bibitem{Srivastava2014}
Nitish Srivastava, Geoffrey Hinton, Alex Krizhevsky, Ilya Sutskever, and Ruslan
  Salakhutdinov.
\newblock {Dropout: A simple way to prevent neural networks from overfitting}.
\newblock {\em The Journal of Machine Learning Research}, 15(1):1929--1958,
  2014.

\bibitem{Bjorne2012}
Jari Bj{\"{o}}rne, Filip Ginter, and Tapio Salakoski.
\newblock {University of Turku in the BioNLP'11 Shared Task}.
\newblock {\em BMC Bioinformatics}, 13(Suppl 11):S4, jun 2012.

\bibitem{Bjorne2014}
Jari Bj{\"{o}}rne.
\newblock {Biomedical event extraction with machine learning}.
\newblock 2014.

\bibitem{Bjorne2015}
Jari Bj{\"{o}}rne and Tapio Salakoski.
\newblock {TEES 2.2: Biomedical Event Extraction for Diverse Corpora.}
\newblock {\em BMC bioinformatics}, 16 Suppl 16(Suppl 16):S4, 2015.

\bibitem{Deleger2016}
Louise Del{\.{e}}ger, Robert Bossy, Estelle Chaix, Mouhamadou Ba, Arnaud
  Ferrė, Philippe Bessieres, and Claire Nėdellec.
\newblock {Overview of the bacteria biotope task at bionlp shared task 2016}.
\newblock In {\em Proceedings of the 4th BioNLP Shared Task Workshop}, pages
  12--22, 2016.

\bibitem{Lever2016}
Jake Lever and Steven J~M Jones.
\newblock {VERSE: Event and relation extraction in the BioNLP 2016 Shared
  Task}.
\newblock In {\em Proceedings of the 4th BioNLP Shared Task Workshop}, pages
  42--49, 2016.

\bibitem{Mehryary2016}
Farrokh Mehryary, Jari Bj{\"{o}}rne, Sampo Pyysalo, Tapio Salakoski, and Filip
  Ginter.
\newblock {Deep learning with minimal training data: TurkuNLP entry in the
  BioNLP shared task 2016}.
\newblock In {\em Proceedings of the 4th BioNLP Shared Task Workshop}, pages
  73--81, 2016.

\bibitem{Li2016}
Honglei Li, Jianhai Zhang, Jian Wang, Hongfei Lin, and Zhihao Yang.
\newblock {DUTIR in BioNLP-ST 2016: utilizing convolutional network and
  distributed representation to extract complicate relations}.
\newblock In {\em Proceedings of the 4th BioNLP Shared Task Workshop}, pages
  93--100, 2016.

\end{thebibliography}





\end{document}